\documentclass[a4paper]{article}
\usepackage{adjustbox}
\usepackage[top=1.in, bottom=1.in, left=1.in, right=1.in]{geometry}
\usepackage{tabls}
\usepackage{cites}
\usepackage{epsf}
\usepackage{appendix}
\usepackage{ragged2e}
\usepackage{mathtools}
\usepackage{mdframed}
\usepackage{caption}
\usepackage{physics}
\usepackage{enumitem}
\usepackage{mathtools}
\usepackage{braket}
\usepackage{booktabs}
\usepackage{tabularx}
\usepackage{makecell}
\usepackage[flushleft]{threeparttable}
\usepackage{algorithm}
\usepackage{algpseudocode}
\usepackage{nomencl} 
\makenomenclature
\usepackage{tikz}
\usepackage{graphicx}
\usepackage{float}
\floatstyle{plaintop}
\restylefloat{table}

\floatstyle{plain}
\restylefloat{figure}
\usepackage{amssymb}
\usepackage{pifont}
\usepackage{amsmath} 

\newcommand{\defeq}{\vcentcolon=}

\definecolor{light-gray}{gray}{0.95}
\newcommand{\code}[1]{\colorbox{light-gray}{\small\color{blue}\textbf{\texttt{#1}}}}

\def\bmheadfont{\reset@font\fontfamily{\rmdefault}\fontsize{10bp}{12bp}\bfseries\selectfont\raggedright\boldmath}%
\newcommand\bmhead{\@startsection{subparagraph}{5}{\z@}%
                                 {6pt \@plus1ex \@minus .2ex}%
                                 {-1em}%
                                 {\bmheadfont}}

\usepackage{url}
\providecommand{\keywords}[1]
{
  \small	
  \textbf{\textit{Keywords---}} #1
}

\title{ Multi-Objective Reinforcement Learning-based Approach for Pressurized Water Reactor Optimization}
\author{  \textbf{Paul Seurin$^1$\footnote{Corresponding author. \textit{E-mail address}: paseurin@mit.edu (P. Seurin)}, Koroush Shirvan$^1$} \\ 
  \\ $^1$Massachusetts Institute of Technology \\ 77 Massachusetts Avenue, Cambridge MA, 02139 \\  \url{paseurin@mit.edu}, \url{kshirvan@mit.edu}}
\begin{document}
\maketitle

\justify 
\begin{abstract}
A novel method, the Pareto Envelope Augmented with Reinforcement Learning (PEARL), has been developed to address the challenges posed by multi-objective problems, particularly in the field of engineering where the evaluation of candidate solutions can be time-consuming. PEARL distinguishes itself from traditional policy-based multi-objective Reinforcement Learning methods by learning a single policy, eliminating the need for multiple neural networks to independently solve simpler sub-problems. Several versions inspired from deep learning and evolutionary techniques have been crafted, catering to both unconstrained and constrained problem domains. Curriculum Learning (CL) is harnessed to effectively manage constraints in these versions. PEARL's performance is first evaluated on classical multi-objective benchmarks. Additionally, it is tested on two practical PWR core Loading Pattern (LP) optimization problems to showcase its real-world applicability. The first problem involves optimizing the Cycle length ($L_C$) and the rod-integrated peaking factor ($F_{\Delta h}$) as the primary objectives, while the second problem incorporates the mean average enrichment as an additional objective. Furthermore, PEARL addresses three types of constraints related to boron concentration ($C_b$), peak pin burnup ($Bu_{max}$), and peak pin power ($F_q$). The results are systematically compared against conventional approaches from stochastic optimization. Notably, PEARL, specifically the PEAL-NdS (\code{crowding}) variant, efficiently uncovers a Pareto front without necessitating additional efforts from the algorithm designer, as opposed to a single optimization with scaled objectives. It also outperforms the classical approach across multiple performance metrics, including the Hyper-volume. 
Future works will encompass a sensitivity analysis of hyper-parameters with statistical analysis to optimize the application of PEARL and extend it to more intricate problems.
\end{abstract} \hspace{10pt}

\begin{mdframed}
\begin{table}[H]   
\nomenclature{RL}{Reinforcement Learning}
\nomenclature{CL}{Curriculum Learning}
\nomenclature{PPO}{Proximal Policy Optimization}
\nomenclature{MDP}{Markov Decision Process}
\nomenclature{CO}{Combinatorial Optimization}
\nomenclature{MOO}{Multi-Objective Optimization}
\nomenclature{SO}{Stochastic Optimization}
\nomenclature{ET}{Evolutionary Techniques}
\nomenclature{NSGA}{Non-dominated Sorting Genetic Algorithm}
\nomenclature{SA}{Simulated Annealing}
\nomenclature{TS}{Tabu Search}
\nomenclature{HV}{Hyper-Volume}
\nomenclature{PEARL}{Pareto Envelope Augmented with Reinforcement Learning}
\nomenclature{FOMs}{Figure Of Merits}
\nomenclature{FA}{Fuel Assembly}
\nomenclature{PWR}{Pressurized Water Reactor}
\nomenclature{$L_C$}{Cycle Length}
\nomenclature{LP}{Loading Pattern}
\nomenclature{$F_{\Delta h}$}{Rod integrated peaking factor}
\nomenclature{EOC}{End Of Cycle}
\nomenclature{$F_q$}{Peak pin power}
\nomenclature{$C_b$}{Boron concentration}
\nomenclature{$Bu_{max}$}{Peak pin burnup}

\printnomenclature
\end{table}
\end{mdframed}


\section{Introduction} 
\label{sec:introduction}

The constrained optimization problem associated with large-scale Pressurized Water Reactors (PWRs), particularly the Loading Pattern (LP) optimization, has been a focal point since the dawn of the commercial nuclear industry. Addressing this challenge requires a reactor designer to select a set of Fuel Assemblies (FAs) for refueling. A reactor typically utilizes a specific number of solid FAs, with two types to choose from: either fresh or burned, ranging from as few as 100 to over 200 FAs. Approximately two-thirds of these FAs are already in the core for a designated period (typically 18 or 36 months), while the remaining FAs are fresh and unused. The selection of fresh FAs involves discrete choices, corresponding to specific input parameters such as the average enrichment and burnable poison loading of the fuel. The resulting fuel arrangement is what we call a LP. This optimization problem falls within the domain of large-scale Combinatorial Optimization (CO) and is characterized by multiple objectives, including economic competitiveness, as well as various physics and operational constraints that limit the number of feasible solutions \cite{seurin2022pwr,seurin2024assessment}.

Previous studies have employed Reinforcement Learning (RL) to optimize the Levelized Cost of Electricity (LCOE) while adhering to safety, regulatory, and operational constraints \cite{seurin2022pwr,seurin2024assessment}. However, these studies focused on a single objective—LCOE—incorporating constraints within an extended objective function penalizing infeasible solutions. As mentioned earlier, the problem of interest is inherently multi-objective and considering multiple objectives simultaneously poses a challenge that must be solved. Nevertheless, the existing RL-based approaches cannot address that. This paper aims to extend the methodologies developed in \cite{seurin2022pwr,seurin2024assessment} to Multi-Objective Optimization (MOO) using RL.

It's worth noting that not only our specific problem but the entire field of nuclear power engineering inherently involves multiple objectives \cite{stewart2021asurvey}. The field is made up of combination of different disciplines of engineering and physics which bring their own unique objectives and constraints. The field of nuclear power engineering has benefited from many advances in optimization algorithms from the computational industry such as the non-dominated sorting algorithm (NSGA-II/III), Pareto Ant colony optimization (P-ACO), and Multi-Objective Cross-Entropy Method (MOOCEM) \cite{schlunz2016acomparative}, Multi-Objective Tabu Search (MOOTS) \cite{mawdsley2022incore} or even Multi-Objective Simulated Annealing-based (MOSA) method \cite{park2009multiobjective}. Nevertheless, these techniques are not widely leveraged in the Nuclear field, which contrasts with others such as in chemical engineering \cite{stewart2021asurvey} or for thermodynamics of heat engines \cite{kumar2016multiobjective}, which are relying more and more on these tools thanks to the increase in computation power \cite{stewart2021asurvey}. Deep reinforcement learning (DRL) based approaches offer novel methods which could be utilized in the context of nuclear power engineering as well.

Deep learning and in particular deep reinforcement learning (DRL) \cite{bertsekas1996neuro} are frameworks that do not work with analytical representation of an objective function but with objects (e.g. function approximator) that are constructed statistically with a large number of data and then mathematically optimized \cite{bengio2018machine}. There are two classical ways to leverage deep learning for optimization: either replacing the function evaluation (e.g., a simulator) by a surrogate, or use it to learn a set of decision rules about the underlying structure of a problem as for RL to generalize from seen to unseen instances of the problems \cite{seurin2022pwr,seurin2024assessment,bengio2018machine}. In the context of large-scale CO, a policy (or sequence of decision) is approximated as a neural network, which weights are trained to generate solution end-to-end of unseen instances of problems such as the Traveling Salesman (TSP) or the Knapsack (KP) very fast \cite{mazyavkina2021reinforcement}. 
A simple forward pass from the trained networks that implement $P(Y|X) =\prod_{k=1}^NP(y_{k+1}|y_1,..,y_k,X_k)$, where $X = (x_1,...,x_N)\in \mathbb{R}^p$ is an input embedding and $Y = (y_1,...,y_N)$ is a permutation of elements, provides the solution. With the concept of efficient generalization in mind, Deep RL (DRL) has demonstrated remarkable prowess in handling classical CO problems, allowing it to generalize effectively from observed instances to previously unseen ones \cite{khalil2017learning}. This feature has also found valuable application in MOO alongside RL. For instance, in \cite{li2020deep}, the TSP is decomposed into N (= 100) scalar objectives obtained from the weighted sum of the different tour costs, each of which is modeled as its neural network. A Neighborhood-based parameter-transfer strategy is utilized to uncover the Pareto front: all sub-problems are solved sequentially and the optimized weights of each neural network are transferred to the next, as neighboring points on the Pareto surface should be close enough, which accelerates the search of new points on the surface. The authors of this study assert that their method surpasses classical heuristic-based MOO methods in terms of computation time, the number of solutions found, and Hyper-Volume (HV). Further instances of studies and techniques for applying MOO with RL, as well as their limitations in the context of our specific problem, are elaborated upon in Section \ref{sec:applicationofRLto}. Notably, a critical limitation, which was emphasized in our previous study \cite{seurin2024assessment}, arises when implementing an end-to-end inference strategy for MOO, particularly when the objective function demands substantial computational time for evaluation \cite{solozabal2020constrained,seurin2024assessment}.  Similar to the single-objective scenario, there was a pressing need to identify alternative strategies for harnessing RL's potential in MOO, specifically within the realm of engineering design.

In this paper, we train a single policy to recover a Pareto front of high-quality solutions. To achieve this, we introduce three novel iterations of an algorithm christened as \textbf{P}areto \textbf{E}nvelope \textbf{A}ugmented with \textbf{R}einforcement \textbf{L}earning (PEARL). The first iteration, denoted as PEARL-envelope (PEARL-e), draws inspiration from deep learning principles. The remaining two iterations employ a ranking system for newly generated points, which serves as a reward mechanism. The first is called PEARL-\code{$\epsilon$}, in which the rank is calculated following the $\epsilon$-indicator strategy in evolutionary multi-objective \cite{lui2021indicator}. The last one, referred to as PEARL-Nondominated Sorting (PEARL-NdS), is rooted in \code{crowding} \cite{deb2002afast} and \code{niching} \cite{jain2014anevolutionaryI,jain2014anevolutionaryII} methodologies. Both iterations PEARL-e and PEARL-NdS have two sub-versions, which will be explicated in Section \ref{sec:derivationofthePareto}.
 
While partial results were previously presented at a conference \cite{seurin2023pareto}, this paper significantly extends the scope of PEARL to address constrained problems, introducing what we term here Constrained-PEARL (C-PEARL). Utilizing the concept of hierarchical networks \cite{ma2019combinatorial}, the problem is divided into distinct two sub-problems. The first sub-problem focuses on identifying feasible solutions, while the second revolves around multi-objective optimization within the feasible space. To accomplish this, we leverage the concept of Curriculum Learning \cite{bengio2009curriculum}, wherein a policy is trained to sequentially tackle the two sub-problems. Initially, it learns to generate feasible solutions, gradually progressing to the more complex task of approximating the optimum Pareto front within the feasible space. \code{CL} is grounded in the idea that learning is more effective when examples are presented in a structured and organized manner, starting with simpler examples and progressing to more intricate ones \cite{bengio2009curriculum}.

The effectiveness and applicability of PEARL are rigorously assessed through a series of comprehensive tests on established classical multi-objective benchmark problems: the dtlzX test suite \cite{deb2002scalable} for the unconstrained cases, and the cxdtlzy \cite{jain2014anevolutionaryII} and ctpx \cite{deb2001constrained} test suite for the constrained cases. Then, we compare their performance on a typical Pressurized Water Reactor (PWR) design \cite{seurin2022pwr,seurin2024assessment} against classical SO-based approaches: NSGA-II \cite{deb2002afast} and -III \cite{jain2014anevolutionaryI,jain2014anevolutionaryII}, Simulated Annealing (SA), and Tabu Search (TS), which are established approaches used for multi-objective optimization\cite{stewart2021asurvey,schlunz2016acomparative,mawdsley2022incore,park2009multiobjective}. These methods have a proven track record in various applications within the nuclear field, including neutron spectrum estimation and advanced reactor optimization \cite{stewart2021asurvey,zeng2020development}. A summary of the RL-based algorithms developed and tested for this work for the unconstrained and constrained cases is given in table \ref{tab:summaryofthealgorithmdev}:
\begin{table}[H]
    \centering
    \caption{Summary of the algorithms developed and tested in this work. '-\code{CL}' in the constrained case means that the constraints are first solved in a single-objective manner and then only the ,ulti-objective version is called.}
    \begin{tabular}{c|c|c|c}
    \hline
    & PEARL-e & PEARL-\code{$\epsilon$} & PEARL-NdS \\
    \hline
   \makecell{Unconstrained \\ variant(s)} & \code{cos},\code{KL-div}& \code{$\epsilon$} & \code{crowding},\code{niching} \\
   
    \makecell{Constrained \\ variant(s)}     &  -\code{CL} & -\code{CL} & -\code{CL},-\code{crowding2}$^{\star}$,-\code{niching2}\\
    \end{tabular}
    \label{tab:summaryofthealgorithmdev}
\end{table}
    \begin{tablenotes}
      \item  \scriptsize$^{\star}$Cases \code{crowding2} and \code{niching2} are similar to the \code{CL} but where the unfeasibility is encapsulated in the ranking strategy (see Section \ref{sec:CurriculumLearningandhierarchical}).
    \end{tablenotes}
    
In summary, this paper makes the following significant contributions:
\begin{enumerate}
      \item Introduction of a pioneering multi-objective policy-based algorithm designed to address both integer and continuous optimization challenges, particularly in the realm of engineering designs.
    \item Pioneering application of multi-objective RL, both in unconstrained and constrained settings, to the core LP problem. This extends the groundwork laid in our previous work on single-objective deep RL for the same problem  \cite{seurin2022pwr,seurin2024assessment,seurin2023pareto}.
    \item Performance comparison against legacy SO-based multi-objective approaches traditionally applied in the field of Nuclear Engineering.
\end{enumerate}

Section \ref{sec:multiobjectivereinforcement} defines the multi-objective framework in the general case (in Section \ref{sec:definitionofmultiobjective}) and within the RL framework (in Section \ref{sec:definition}) and provides several applications, Section \ref{sec:derivationofthePareto} unveils the derivation of our PEARL algorithms for the unconstrained (section \ref{sec:unconstrainedPEARL}) and constrained PEARL (section \ref{sec:constrainedPEARLcPEARL}), and Section \ref{sec:Designofexperiments} describes the different problems PEARL will be applied to and compared: the benchmarks in Section \ref{sec:classicallbenchmarks} and the pwr designs in Section \ref{sec:pwrmultiobjectiveproblems}. The results and analysis are given in the Section \ref{sec:results}, while the concluding remarks are given in Section \ref{sec:concludingremarks}.

\section{Multi-Objective Reinforcement Learning (RL)}
\label{sec:multiobjectivereinforcement}
\subsection{Definition of Multi-Objective Optimization (MOO)}
\label{sec:definitionofmultiobjective}

Optimization is a notion that encapsulates a large spectrum of mathematical tools, algorithms, and search procedures and underlies every engineering decision-making process. The goal of optimization for a designer is to find the best (or optimum) solution based on a pre-defined set of design goals (e.g., economics, performance, reliability, ...) under a set of design constraints (e.g., manufacturing, safety, materials, ...) that can be conflicting, by perturbing a set of decision variables (e.g., amount of input materials, the geometry of the structure...). 
Every engineering decision-making process relies on solving an optimization problem, which can be formulated without loss of generality as follows:
\begin{align}
\begin{split}
    &\min_{x} f(x) = [f_1(x),f_2(x),....,f_F(x)]^T \\
    &\text{s.t.} \quad g_i(x) \le 0 \quad \forall i\in \{1,...,N\} \\
     & h_j(x) = 0 \quad \forall j\in \{1,...,M\}\\
     & x = [x_1,x_2,...,x_K]^T \in \Gamma \subset \mathbb{R}^K
\end{split}
\end{align}
\label{eq:fObj}
\noindent
where $(f_i)_{i \in \{1,...,F\}}:\Gamma \rightarrow \mathbb{R}$ is a set of F design objectives, $(g_i)_{i \in \{1,...,N\}}$ is a set of N design inequality constraints, $(h_j)_{j \in \{1,...,M\}}$ is a set of M design equality constraints, $x\in \Gamma \subset \mathbb{R}^p$ are decision variables. If F is equal to 1 the problem is coined single objective. 
 In a MOO problem, a designer is trying to maximize a set of usually conflicting objectives rather than only one. One method to solve this problem is to turn it into multiple single-objective ones and is called the weighted sum or multiplicative penalty \cite{stewart2021asurvey,schlunz2016acomparative}. There, all objectives are encapsulated in a single pseudo-objective by weighting each objective depending on the importance of each with weights $w = (w_1,...,w_F)$. Let $w\in \mathbb{R}^F$ a well defined \textbf{preference vector}, the problem becomes (we omit the constraints for simplicity): $\min_{x} \sum_{i = 1}^F w_if_i(x)$ such that $\sum_{i=1}^F w_i = 1$. However, to diversify the solution's set, a user has to run the optimizer multiple times with different $w$ to obtain the desired level of trade-offs. This can become time-consuming and cannot capture concave Pareto fronts \cite{chen2020combining}. Methods offering a wealth of solutions in one run would provide more flexibility \cite{schlunz2016acomparative} and help the designer makes quantitative evidence-based decisions \cite{parks1996multiobjective}. 
 The framework of MOO offers the possibility to characterize such a set with the notion of Pareto front \cite{stewart2021asurvey}. The solutions which belong to the Pareto front are said to be non-dominated. Let $\mathbb{O} \subset \Gamma$ be the set of feasible solutions and $(f_i)_{i \in \{1,...,F\}}$ a set of $F$ objectives. Without loss of generality, $y$ is said to be \textit{dominated} by $x$ (and we note \textbf{$x \succ y$}) if $\forall p \in \{1,...,F\}, \quad f_p(x) \le f_p(y)$ and $f_k(x) < f_k(y)$ for at least one $k\in \{1,...,F\}$. Therefore, departing from any solution on this front would deteriorate one or another objective. The Pareto-optimal set $\mathcal{P}^{\star}$ and front $\mathcal{PF}^{\star}$  can be denoted as follows \cite{riquelme2015performance}:

 \begin{equation}
     \mathcal{P}^{\star} \defeq \{ \tilde{x} \in \mathbb{O} | \nexists x, x \succ \tilde{x} \} \quad \mathcal{PF}^{\star} \defeq \{(f_i)_{i \in \{1,...,F\}}(\tilde{x})|\tilde{x} \in \mathcal{P}^{\star}\}
 \end{equation}
 
When $x$ and $y$ do not dominate each other we say that they are \textit{Pareto equivalent}. Note that if $x \in \mathbb{O}$ and $y \notin \mathbb{O}$, then $x \succ y$. If $x,y \notin \mathbb{O}$, $x \succ y$, if $CV(x) < CV(y)$, where the Constraint Violation ($CV$) is a function of the constraints $CV(x) = CV(g_1(x),..,g_N(x),h_1(x),...,h_M(x))$ \cite{jain2014anevolutionaryII}.  For many real world application the front is not known because it would be computationally infeasible to find it. Therefore, the designer must fall back on approximation of these sets which we denote $\tilde{\mathcal{P}^{\star}}$, $\tilde{\mathcal{PF}^{\star}}$ obtained for instance by considering all the non-dominated solutions from an ensemble of various runs \cite{schlunz2016acomparative,mawdsley2022incore,wang2022investigating,qi2022aqlearning}. The main goals of a MOO framework are then (1) to find the best approximation of the set of Pareto-optimal solutions (sometimes called \textbf{convergence criterion} \cite{riquelme2015performance,yang2019amanyobjective}), (2) to provide an extensive amount of trade-offs  (also known as the \textbf{diversity criterion} \cite{riquelme2015performance,yang2019amanyobjective}), in which the solutions are diverse and dispersed and (3) to provide a large number of solutions (also known as the \textbf{cardinality} criterion) \cite{riquelme2015performance}.

\subsection{Definition of RL within the MOO context}
\label{sec:definition}

RL is the subclass of machine learning algorithm which deals with decision-making. In RL, one or several agents in parallel are trained to take a sequence of actions $a \in \mathcal{A}$ from observable states $s \in \mathcal{S}$ following a policy: $a \sim \pi(\cdot,s;\theta_k)$, where $\theta_k$ is a learnable parameter to improve this sequence. When the learnable parameters are part of a neural network, it is classically called Deep Reinforcement Learning (DRL). We loosely utilize the term RL for both cases. A policy $\pi = (\pi_1,...)$ is a sequence of decision rules, strategies, or plans, where $\pi_t : \mathcal{S} -> \Delta(\mathcal{A})$, at time step $t$, represents the probability of taking action $a \in \mathcal{A}$ being in state $s \in \mathcal{S}$. It specifies which action to take when in state $s$. If the policy is deterministic then $\Delta(\mathcal{A}) = \mathcal{A}$ otherwise $\Delta(\mathcal{A}) \subset [0,1]$. When the decision does not change over time (i.e., $\pi = (\pi_1,\pi_1,...,\pi_1)$), it is called a \textit{stationary} policy, otherwise it is called \textit{non-stationary}. Each action $a \in \mathcal{A}$ taken after observing a state $s \in \mathcal{S}$ generates a reward $r(s,a) \in \mathbb{R}$ or $r(s,a) \in \mathbb{R}^F$ in the case of multi-objective RL where F is the number of objectives \cite{yang2019ageneralized,mossalam2016multiobjective}.
 A rigorous definition of RL and Markov Decision Process (MDP) applied to PWR optimization is given in \cite{seurin2022pwr,seurin2024assessment}. There, the core LP optimization problem was solved for the first time with Proximal Policy Optimization (PPO), which is an RL algorithm. However, it was solved with a single objective in mind, namely reducing the Levelized Cost of Electricity (LCOE), while satisfying a set of safety and operational constraints. For multi-objective RL the definition differs slightly. For the latter, the solution to the problem is a set of policies called the coverage set (CS), which are all optimum for at least one problem defined as follows:
 
 Let $f:\mathbb{R}^F \times \Omega\rightarrow \mathbb{R}$ a function over vectorized value functions $V^{\pi}$ and $w = \{w_1,w_2,...,w_F\} \in \Omega \subset [0,1]^{F}$ a preference vector representing the importance of each sub-objective or cumulative utility, and $\Omega$ the space of preferences.  $f(V^{\pi},w)$ is called a \textit{utility function}. For instance, it is equal to $w \cdot V^{\pi}$ if $f$ is linear and we can call CS the Convex CS (CCS) \cite{yang2019ageneralized,mossalam2016multiobjective}. For N preference vectors (adapted from \cite{ruchte2021efficient} in the context of multi-task learning), the problem becomes to find $\pi(\cdot|s;\theta^{\star})$ such that :
\begin{align}
\begin{split}
    & \theta^{\star} \in arg\max_{\theta} \quad \mathbb{E}_{\mathcal{H}}\{\Sigma_{t = 0}^{\tau} \gamma^t r_t^1(s_t,a_t),\Sigma_{t = 0}^{\tau} \gamma^t r_t^2(s_t,a_t),...,\Sigma_{t = 0}^{\tau} \gamma^t r_t^N(s_t,a_t)\} \\ 
    & = \quad arg\max_{\theta} \{ f(w_1,V^{\pi_{\theta}}),f(w_2,V^{\pi_{\theta}}),...,f(w_{N},V^{\pi_{\theta}}))\}\\ 
\end{split}
\end{align}
The coverage set is then defined as the set that contains all the policies that maximize the utility $f$ for a subset of preference vectors:
\begin{equation}
CS := \{ V^{\pi^{i}_{\theta}} | \exists w^{\star} \in \Omega \quad \text{s.t.} \quad f(V^{\pi^{i}_{\theta}},w^{\star}) \ge f(V^{\pi^{j}_{\theta}},w^{\star}), \forall \pi^{j}_{\theta} \in \Pi\}
\end{equation}

where $\Pi$ is the manifold of smoothly parametrized policies.

\subsection{Application and limitations of RL to MOO}
\label{sec:applicationofRLto}

Multi-objective problems are commonplace in RL and often entail striking a delicate balance between various objectives. Consider, for instance, the task of teaching a robot to run, where a policy must effectively manage the trade-offs between running speed and energy efficiency \cite{xu2020predictionguided}. In the introduction, we alluded to a method that relies on training multiple policies to solve a problem end-to-end. However, many other techniques exist that leverage the swath of possibilities these new advancements in AI offer. Overall, three other methods were classically utilized: encapsulating the multi-objective nature of the reward (e.g., via HV) in the Q-update \cite{moffaert2014multiobjective}, sampling the preference vector before evaluating a solution found \cite{yang2019ageneralized}, or turning the problem into a single-objective optimization via different techniques (e.g., linear scalarization, minimum, $L_p$-, and $L_{\infty}$-metrics compared to a known set of Pareto equivalent solutions \cite{xu2021multiobjective}) also called min-max method in \cite{stewart2021asurvey}. 

One additional approach is to use a meta-policy, which is a general policy that can be tuned to a particular trade-off, but it may not always be optimal \cite{xu2020predictionguided}. Another approach is envelope Q-learning, which is a multi-objective version of the Q-learning algorithm \cite{yang2019ageneralized}. The authors attempted to learn a single policy and generalize it to any preference vector by splitting between a learning and an adaptation phase. For each update, not only do they sample the transitions to update the network parameters, but they also sample a set of preference vectors to then maximize over the convex hull of the solution frontier with a new type of bellman optimality operator and composite loss. The method demonstrated high performances against classical multi-objective RL techniques. \cite{chen2020combining} also leveraged the neighboring idea, where the authors created an algorithm to learn multiple policies in synergy. Each policy is assigned a scalarized objective ($w\cdot V^{\pi}$ where $w$ is a set of weights and $V^{\pi}$ is a value vector) which is learned with the off-policy RL algorithm Soft Actor Critic (SAC) method. The networks share a lower layer which reduces the number of parameters learned and plays the role of \textit{representation learning} for more robust learning. A multi-channel replay buffer is maintained to improve Sample Efficiency (SE), in which the experience generated by each policy may or may not be leveraged by the other policies during the network updates. They then use a multi-objective Covariance Matrix Adaptation Evolution Strategy (MO-CMA-ES) to fine-tune policy-independent parameters to improve the Pareto front.

However, these approaches have limitations. The end-to-end inference strategy in \cite{li2020deep} is not implementable when the objective function is time-consuming to evaluate \cite{seurin2024assessment,solozabal2020constrained}, which is the case for the problem at hand \cite{seurin2022pwr,seurin2024assessment}. The applications in \cite{yang2019ageneralized} did not cover combinatorial optimization, which is the case for this work \cite{seurin2022pwr,seurin2024assessment}. Moreover, we argued in this past work that Q-based approaches under-performed against policy-based ones (as used in \cite{yang2019ageneralized,xu2021multiobjective,moffaert2014multiobjective}). Additionally, the technique requiring multiple policies (in \cite{chen2020combining,li2020deep,xu2020predictionguided}) induces computational burden because the larger number of parameters to train can cause slowness \cite{chen2020combining}, and separating networks is not adapted to GPUs usually used when training environments with image inputs \cite{chen2022redeeming}. It is also difficult to implement this tactic in stable-baselines \cite{stable-baselines}, our RL algorithm package. Lastly, the scalarization-based metrics (utilized in the approaches \cite{xu2021multiobjective,wang2022investigating}), may be limited, as two Pareto equivalent solutions may yield different scalarized values, and therefore an agent will preferentially choose only one of the two. 
 
Hence, the primary objective of this work is to develop a reward mechanism for a single policy that avoids favoring specific Pareto-equivalent points and instead achieves a well-distributed Pareto envelope in a single run, without the need for additional trainable parameters. To accomplish this, we introduce the Pareto Envelope Augmented with Reinforcement Learning (PEARL) algorithm, which involves training a single policy to address a wide range of Multi-Objective Optimization (MOO) problems within the constraints of engineering design. Section \ref{sec:derivationofthePareto} provides the derivation of the PEARL algorithm.

\section{Derivation of the Pareto Envelope Augmented with Reinforcement Learning (PEARL)}
\label{sec:derivationofthePareto}

\subsection{Unconstrained PEARL}
\label{sec:unconstrainedPEARL}
In \cite{seurin2022pwr,seurin2024assessment} for the single-objective case, we have empirically shown that the optimum way to apply RL for solving an optimization problems is to transform it into what we could call an \textit{automatic supervised setting}. While generating one LP per action with horizon (i.e, episode) length equal to 1, we have a value function $V(s_0;\theta_{V})$ which is trained to predict the reward of each LP with supervised samples generated by a learning policy $\pi(\cdot;\theta_{\pi})$, where $\theta_{V}$ and $\theta_{\pi}$ are learnable parameters. We have also shown that the policy converges over time to a solution equal or close to the best one ever found. Because of the closeness of the previous approach to supervised learning, a natural extension of it to MOO must draw inspiration from multi-objective supervised learning, in particular Multi-Task Learning (MTL) \cite{mahapatra2020multitask}. For instance in \cite{mahapatra2020multitask} and \cite{ruchte2021efficient}, a preference vector $w$ is sampled at the beginning of the algorithm and before each gradient update, respectively, to form a loss function to train the learning model that encapsulates each objective:  Let $(J_i)_{i\in \{1,...,F\}}$ a set of F losses to jointly minimize, the resulting composite loss is equal to $w^TJ=\sum_{i=1}^{F}w_iJ_i$ (and the model is conditioned on the $w$ to generalize to any $w$). However, a linear scalarization cannot capture the concave part of the Pareto envelope. Therefore, they all added a non-uniformity factor $\mu_u$ that measures the deviation between the solution found and the randomly drawn preference vectors. As such, it also helps them to obtain a dispersed Pareto front. The difference is in the formulation of this factor. The authors in  \cite{mahapatra2020multitask} proposes the formulation in equation \ref{eq:mahapatranonuniformity}:
\begin{equation}
\mu_u(s_t,a_t) = \sum_{i = 1}^F \tilde{r_i(s_t,a_t)} \log{\frac{\tilde{r_i(s_t,a_t)}}{1/F}} = KL(\tilde{r(s_t,a_t)|\frac{1}{F}})
\label{eq:mahapatranonuniformity}
\end{equation}
where $KL(p|q) = \int_{x \in X} p(x) log(\frac{p(x)}{q(x)})dx$, $\tilde{r_i(s_t,a_t)} = \frac{w_jr_i}{\sum_kw_kr_k}$. $\mu_u$ estimates the extent of which $\tilde{r}$ is non-uniform. Alternatively, in \cite{ruchte2021efficient}, the factor $\mu_u$ is equal to $ - \lambda \frac{w^TJ(x,y,r;\theta)}{||w||\cdot||J(x,y,r;\theta)||}$ (i.e., the cosine similarity), where $\lambda$ is a weighting coefficient to tune. 
 
 These two approaches motivated us to use a non-uniformity penalty term and trace vectors in the objective space to ensure diversity of the Pareto envelope. Hence, we designed a reward augmented with the preference vector $w$ and a sampling procedure which solves:
\begin{equation}
    \theta^{\star} \in arg\max_{\theta} \mathbb{E}_{\substack{\sim \mathcal{H} \\ w\sim \mathcal{D}_{w}}}(\Sigma_{t = 0}^{\tau} \gamma^t (w^Tr(s_t,a_t)+\lambda \mu(w,r(s_t,a_t)))
\end{equation}
where $ \mu(w,r(s_t,a_t))$ is the non-uniformity factor equal to $\frac{w^Tr(s_t,a_t)}{||w||\cdot||r(s_t,a_t)||}$ or $ KL(\tilde{r(s_t,a_t)|\frac{1}{F}})$, and $\mathcal{D}_{w}$ is the preference sampling distribution, chosen to be the Dirichlet as in \cite{ruchte2021efficient}. The advantage of using the Dirichlet distribution is its flexibility in taking various shapes based on the value of $\alpha$. This is particularly useful since the shape of the Pareto front is not known a priori, and the ability to adjust the location of the rays in the objective space is crucial. Within the PPO paradigm, we sample $w$ every \code{n\_steps}. Additionally, because the magnitude of different objectives can differ drastically, We also provide the opportunity to normalize the reward similar to Indicator-based MO-EA (IBEA) \cite{Zitzler2004IndicatorBasedSI} via a boolean \code{normalized\_obj}. This method is labeled \textbf{\textit{PEARL-e}}. When the non-uniformity is based on \cite{mahapatra2020multitask} and \cite{ruchte2021efficient}, we note PEARL-e (\code{KL-div}) and PEARL-e (\code{cos}), respectively.

Our second set of methods is inspired by SO and in particular Evolutionary Techniques (ET). We will leverage ET formulations to shape the reward which serves as a measure of the quality of the information provided by a newly generated solution on the existing approximation of the Pareto front. One such reward can be indicator-based. The additive $\epsilon-$indicator $I_{\epsilon}^+$ (e.g., IBEA \cite{yang2019amanyobjective}) is the method of choice due to its popularity \cite{riquelme2015performance} and the fact that it is inexpensive to compute \cite{yang2019amanyobjective}. Let a population of known pareto equivalent solution $\mathcal{P}$ and $F_{I_{\epsilon}^{+}}(x_t)$ the fitness of the new solution generated $x_t$ at time step t.  $F_{I_{\epsilon}^{+}}(x_t)$ represents the loss of quality by adding $x_t$ to the population (see \cite{yang2019amanyobjective} for the exact definition). In our case, the population is a buffer of non-dominated solutions encountered during the search because, unlike the original ET-based paper \cite{jain2014anevolutionaryI}, we do not have a population to play the role of the optimum front. Our choice draws inspiration from MOSA \cite{park2009multiobjective} and or Pareto Evelope-based Selection Algorithm (PESA) \cite{konak2006multiobjective}, where a reference buffer was used to overcome this issue. First, any solution $(y_i,r_i)_{i \in \{1,...,k\}}$ dominated by the new solution $(x_t,r_t)$ is removed, while if $x_t$ is dominated by any of the existing solutions, it is not kept and assigned a negative reward equal to the maximum size of the buffer of Pareto-equivalent solutions \code{$\kappa$} specified by the user (equivalent to the population size in IBEA). Otherwise, we measure $F_{I_{\epsilon}^{+}}(y_i)$ $\forall y_i \in \mathcal{P}$ and rank $x$ based on $F_{I_{\epsilon}^{+}}(x_t)$. This method is labeled \textbf{\textit{PEARL-\code{$\epsilon$}}}.
 Alternatively, we could assign a reward based on the distribution of solution on the pareto front, which draw inspiration from the density-based methods in multi-objective GA, namely NSGA-II \cite{deb2002afast} and NSGA-III \cite{jain2014anevolutionaryI,jain2014anevolutionaryII}. Therefore, we modified the NSGA-II/III ranking assignment by also applying a density-based ranking procedure to a buffer of Pareto equivalent solutions and a new solution $x$. This method is labeled \textbf{\textit{PEARL-Nondominated Sorting (PEARL-NdS)}}. When the method is based on NSGA-II and NSGA-III, we note PEARL-NdS (\code{crowding}) and PEARL-NdS (\code{niching}), respectively.
 
 The overall procedure with PEARL-e, PEARL-\code{$\epsilon$}, and PEARL-NdS is summarized in algorithm
 \ref{alg:PEARLepeusocode}:

\begin{algorithm}[H]
 \small
   \caption{Pseudocode for PEARL algorithms at time step t}
   \label{alg:PEARLepeusocode}
   \begin{algorithmic}[1]
      \State \textbf{Input:}  maximum size of the buffer \code{$\kappa$}, buffer $E_b^{t - 1} = \{r_b^1,r_b^2,...,r_b^k\}$,  $w \sim\mathcal{D}_{w}(\alpha)$ a preference vector, the type of non-uniformity $\mu(w,\cdot)$, \code{selection} 
 \State \textbf{Output:}  Reward assigned to the new solution generated $\tilde{r_t}$ and Pareto front $E_b^{t}$ at time step t 
 \If{PEARL-e}
              \State \textbullet $w$ is added to each state and in particular the initial $s_0$ at the beginning of an episode
              \EndIf
                    \State \textbullet An action is sampled from the current policy $a_p \in \mathcal{A} \sim \pi_{\theta_t}$
                    \State \textbullet Action $a_p$ is  evaluated by the environment, a new state $s_p \in \mathcal{S}$ is observed, and a reward $r_p \in \mathbb{R}^F$ is returned
                    \State \textbullet Obtain $\mathcal{PF}^t$ the Pareto front of $E_b^{t} = \{r_b^1,r_b^2,...,r_b^k,r_p\}$
                    \If{PEARL-e}
                    \State The envelope reward is evaluated as $\tilde{r_t} = \max_{w_j\in w} \{w_j^Tr_p + \lambda \mu_u(w,r_p) \}$
                   \EndIf
                   \If{PEARL-\code{$\epsilon$} or PEARL-NdS}
                      \If{$r_p \in \mathcal{PF}^t$}
                         \State \textbullet Evaluate the indicator value for each element of the current Pareto front $F_{I_{\epsilon}^{+}}(r_p)$ and rank each one in descending order or apply a diversity-based technique to obtain its rank based on \code{selection}
                       \State \textbullet  $\tilde{r_t} = - rank(r_p)$
                      \Else
                        \State \textbullet Set $\tilde{r_t} = -$  \code{$\kappa$}
                        \EndIf
                   \EndIf
                   
                    \State \textbullet Set $E_b^{t} = \mathcal{PF}^t[:\kappa]$
                     \State \textbullet return $\tilde{r_t}$,$E_b^{t}$
\end{algorithmic}
\end{algorithm}

Nonetheless, it's important to recognize that virtually all real-world applications come with constraints that must be taken into account. Therefore, it becomes imperative for us to develop a methodology that seamlessly integrates constraints into our framework. We have termed this integration of such as methodology to the unconstrained PEARL, Constrained-PEARL (C-PEARL), which extends the versions described earlier to satisfy constraints. A comprehensive description is provided in Section \ref{sec:constrainedPEARLcPEARL}.

\subsection{Constrained PEARL (C-PEARL)}
\label{sec:constrainedPEARLcPEARL}
Section \ref{sec:ConstraintshandlinginReinforcement} summarizes the classical approaches to deal with constraints in RL and Section \ref{sec:CurriculumLearningandhierarchical} describes our proposed methodology C-PEARL.

\subsection{Constraints handling in Reinforcement Learning (RL)}
\label{sec:ConstraintshandlinginReinforcement}
The application of RL to address constraints within optimization problems remains a relatively unexplored area \cite{zhang2020first,solozabal2020constrained}. To grasp the distinctions between constrained-RL and traditional RL methods, it becomes essential to extend the conventional formulation of a MDP into a more comprehensive framework known as a Constrained-MDP (CMDP). Within a CMDP, the existing policies are augmented with constraints, thereby reducing the set of permissible policies \cite{zhang2020first}. This extension allows for a more nuanced and adaptable approach when dealing with optimization problems that involve various constraints: $\Pi_C = \{\pi \in \Pi: J_{C_i}(\pi_{\theta}) \le c_i, \quad i \in \{1,...,M\} \}$ where $J_{C_i} = \mathbb{E}_{\pi_{\theta}}(\sum_t^{\tau} \gamma^tc_i(s_t,a_t))$, $\Pi$, and $c_i(s,a): \mathcal{S}\times \mathcal{A} \rightarrow \mathbb{R}$ is the cost (in contrast to reward) incurred by taking action $a \in \mathcal{A}$ being in state $s\in \mathcal{S}$ (e.g., a robot actuating an arm too often would result in mechanical fatigue and potential break down). The simplified formulation of the CMDP is then:
\begin{align}
\begin{split}
    &\max_{\theta} J_R(\pi_{\theta}) = \mathbb{E}_{\pi_{\theta}}(\sum_t^{\tau} \gamma^tr_i(s_t,a_t)) \\
    & s.t. \quad a_t \sim \pi_{\theta}(\cdot |s_t) \\
    & \quad J_{C_i}(\pi_{\theta}) = \mathbb{E}_{\pi_{\theta}}(\sum_t^{\tau} \gamma^tc_i(s_t,a_t)) \le c_i
\end{split}
\end{align}
where $J_R(\pi_{\theta})$ is the classical exponential average cumulative reward of the original MDP \cite{seurin2024assessment}. The concept of addressing safety concerns in RL, often referred to as "RL-safety," has gained significant traction in recent years. One approach to handling such constraints is to incorporate them directly into the RL algorithm updates. Notable methods in this category include Constrained Policy Optimization (CPO) \cite{achiam2017constrained} and its first-order simplification proposed by \cite{zhang2020first}, which are extension of the classical Trust-Region Policy Optimization (TRPO) algorithm \cite{schulman2017trust}. \cite{zhang2020first} solves the system of equation in a two-step manner in which a primal-dual approach is used to solve the constrained MDP in the non-parametrized policy space, which is then projected back in the parametrized policy space leveraging a minimization of a Kullback-Leibler (KL) divergence solved by stochastic gradient descent. This method has been successfully applied to train agents, such as a robot constrained by a speed limit and an agent tasked with navigating along a circular path while staying within predefined safety boundaries. In \cite{seurin2024assessment}, we argued that PPO outperformed TRPO-based approaches on our problem. Consequently, there was no intent to explore a new set of hyper-parameters for the TRPO-based approach, given the success observed with PPO.

In this work, we are usually seeking for a set of solutions, where not only the sum but the maximum of the constraints over all solutions generated must be below a threshold $J_{C_i}(\pi_\theta) = max_t c_i(s_t,a_t) \le b_i$. Typically, the design generated is unsafe due to one or more physical parameters. This type of constraints are referred to as \textit{instantaneous} \cite{liu2021policy}. There, we can distinguish \textit{implicit} and \textit{explicit} constraints. The latter are constraints that can be dealt with maskable actions, random replacement, or by construction of the CMDP. This is the case for most combinatorial optimization problem (e.g., in the Traveling Salesman Problem (TSP), one episode terminates when each city has been selected to generate a valid tour \cite{kotary2021endtoend,liu2021policy}). For the PWR problem in \cite{seurin2022pwr,seurin2024assessment,seurin2023pareto}, each fuel pattern has exactly the same number of FAs by construction, namely 89. In contrast, \textit{implicit} constraints cannot be solved or a-priori be known in advance, which is the case for most real-world application \cite{seurin2024assessment,ma2019combinatorial,liu2021policy}. To deal with it rather, the in-feasibility that cannot be embedded in the generation of solutions incur a penalty in the reward directly $J_{R_{\lambda}}(\pi_{\theta}) = J_R(\pi_{\theta}) - \sum_i \lambda_i J_{C_i}(\pi_{\theta})$ where $J_{R_{\lambda}}$ is the dual function and $(\lambda_i)_{i \in \{1,...,M\}}$ are $M$ Lagrangian multiplicative coefficients corresponding to $M$ constraints \cite{solozabal2020constrained,ma2019combinatorial}, which is often called \textit{Lagrangian Relaxation} or \textit{augmented Lagrangian}.

However, a particular attention must be geared towards the shape of the reward as infeasible solutions cannot be uncovered if the reward is flat and the policy never learns. This was our method of choice in \cite{seurin2022pwr,seurin2024assessment}, where we factored the constraints in the reward by penalizing the solution with how far it is from the feasible space with a pseudo-objective explicated in equation \ref{eq:NucCoreObj}:
\begin{equation}
    \max_{x} f(x) = - f_c(x) - \sum_{i \in C} \gamma_i \Phi(x_i) + 1 \times \delta_{\forall i, x_i \le c_i}
    \label{eq:NucCoreObj}
\end{equation}
where $f_c(x)$ is the LCOE of candidate $x$, $C = \{ (c_{i})_{i \in \{1,..,7\}} \}$ is the set of constraints, and $(\gamma_{i})_{i \in \{1,..,7\}}$ is a set of weights attributed to each constraint. If $ (x_{i})_{i \in \{1,..,7\}}$ are the values reached for the candidate $x$ for the corresponding constraints $(c_{i})_{i \in \{1,..,7\}}$, $\Phi(x_i)$ is equal to $ \delta_{x_i \le c_i} (\frac{x_i - c_i}{c_i})^2$, where $\delta$ is the Kronecker delta function. This approach works well for a single-objective \cite{seurin2022pwr,seurin2024assessment}. However, in the case of PEARL, we need to follow a buffer of pareto equivalent solutions. It is likely that two solutions could be pareto equivalent in the full space, but one solution is feasible while the other one is not, therefore the former should be preferred. The procedures to deal with this challenge will be elaborated on in Section \ref{sec:CurriculumLearningandhierarchical}.


\subsection{Curriculum Learning and hierarchical approach for Constrained-PEARL (C-PEARL)}
\label{sec:CurriculumLearningandhierarchical}

The concept of segregating constraints from the primary objective has been explored in the context of single-objective optimization for Boiling Water Reactor (BWR) fuel bundle optimization  \cite{radaideh2021physics,radaideh2021large}. In this approach, constraints are systematically addressed in a stepwise manner, and only after they are all satisfied are solutions optimized for the primary objective. This approach bears resemblance to Curriculum Learning (CL) \cite{bengio2009curriculum}, a technique particularly effective in locating local optima within non-convex criteria. In CL, the learning process begins with easy examples and gradually introduces more complex ones that illustrate advanced tasks, accelerating the search and convergence towards better local optima. In \cite{bengio2009curriculum} for instance, they managed to reduce the classification error of a perceptron by initially presenting less noisy (i.e., \textit{easy}) examples to then more noisy (i.e., \textit{complex}) ones. The essence of CL lies in presenting examples in an organized and ordered manner to enable the learning architecture to expand in complexity, with weights converging towards attraction basins, akin to students progressing through a curriculum. The CL technique aligns well with online training, making it compatible with the RL framework. Initially introduced as a regularization method in supervised learning \cite{bengio2009curriculum}, CL has demonstrated numerous successes in RL \cite{rusu2016progressive}.

Seeking to harness a different form of knowledge transfer in the realm of constrained optimization, the authors of \cite{ma2019combinatorial} proposed an alternative solution involving a hierarchical Graph Pointer Network (GPN) to address the Traveling Salesman Problem with Time Windows (TSPTW). In this approach, the problem is decomposed into subproblems (or layers), each learned with different policies. For example, the lowest layer is trained with rewards that encourage feasible solutions, while the highest layer focuses solely on optimizing the objective.

Drawing inspiration from Curriculum Learning (CL) and the sub-problem decomposition seen in hierarchical approaches, our approach involves training a policy network initially to identify feasible solutions. Subsequently, we focus on optimizing the Pareto front within the feasible space, applying the standard PEARL method as described in Section \ref{sec:derivationofthePareto}. Notably, we utilize the same neural networks for all policies and transfer the learned weights from one problem to the next. In this initial experiment of our approach, we do not employ progressive or hierarchical networks as discussed in \cite{rusu2016progressive,ma2019combinatorial}. On top of that, for our application, we found that solving all constraints at once yield to find solutions much faster, as if solving only one constraint at a time drags the search in a bad local optima.

We are faced with two distinct sub-problems in our approach: one entails resolving all constraints, while the other involves constructing a Pareto front. This separation has proven to be effective in practical applications in literature. Many existing methods for multi-objective constrained optimization, rooted in SO approaches, employ a cumulative penalty function to distinguish between feasible and infeasible solutions \cite{jain2014anevolutionaryII,park2009multiobjective}. These methods are referred to as "Constraint Violation" (CV)-based techniques, where the sole magnitude of the constraint violation, denoted as $CV(x)$, defines the quality of a solution if it is non-zero. For instance, to handle constraint in NSGA-III \cite{jain2014anevolutionaryII}, the authors extended the concept of dominance to encompass infeasible solutions by introducing a constraint violation term, $CV(x) = \sum_i C_i(x)$. For a solution $x$, $CV(x)$ computes its distance from the feasible space, yielding a value of 0 for feasible solutions.

Instead of utilizing the magnitude of $CV$ as a reward, an alternative approach is to employ it for ranking solutions, akin to the process of discovering the Pareto front in unconstrained scenarios. This extension of the notion of dominance to unconstrained cases is explained in the introduction. We designate this approach as \code{crowding2} and \code{niching2}, if the ranking is based on \code{crowding} or \code{niching}, respectively. In the reminder of the paper, the \code{CL} for the former will be omitted because they only have a constrained version, while it will be reminded for the latter because they both have an unconstrained version. The former method is magnitude-free, making it suitable for problems where the reward structure is unknown, and practitioners prefer not to alter its formulation. Additionally, this approach alleviates the need for additional hyper-parameter tuning. The pseudocode for C-PEARL is given in in algorithm \ref{alg:cPEARLpeusocode}:
\begin{algorithm}[H]
 \small
   \caption{Pseudocode for C-PEARL algorithms at time step t}
   \label{alg:cPEARLpeusocode}
   \begin{algorithmic}[1]
      \State \textbf{Input:}  maximum size of the buffer \code{$\kappa$}, buffer $E_b^{t - 1} = \{r_b^1,r_b^2,...,r_b^k\}$, \code{selection}, bonus \code{M}
 \State \textbf{Output:}  Reward assigned to the new solution generated $\tilde{r_t}$ and Pareto front $E_b^{t}$ at time step t 
                    \State \textbullet An action is sampled from the current policy $a_p \in \mathcal{A} \sim \pi_{\theta_k}$ a new state $s_p \in \mathcal{S}$ is observed, and a reward $r_p \in \mathbb{R}^F$ is returned
                    \If{\code{niching2} or \code{crowding2} in \code{selection}}
                     \State \textbullet Obtain the Pareto front of $E_b^{t} = \{r_b^1,r_b^2,...,r_b^k,r_p\}$ \textit{with feasibility in dominated criterion}, $\mathcal{PF}^t$
                      \If{$r_t \in \mathcal{PF}^t$}
                     \State \textbullet  $\tilde{r_t} = - rank(r_t)$
                     \Else
                     \State \textbullet  $\tilde{r_t} = -$ \code{$\kappa$}
                     \EndIf
                      \State \textbullet Set $E_b^{t} = \mathcal{PF}^t[;\kappa]$
                    \Else
                    \If{All constraints are satisfied}
                          \State \textbullet Obtain the Pareto front of $E_b^{t} = \{r_b^1,r_b^2,...,r_b^k,r_p\}$,  $\mathcal{PF}^t$
                       \State \textbullet  \[\tilde{r_t} = PEARL(r_p)\]
                        \State \textbullet Set $E_b^{t} = \mathcal{PF}^t[;\kappa]$
           
                     \Else
                   \State \textbullet   \[\tilde{r_t} =  - \sum_{i \in C} \gamma_i \Phi(x_i) - \code{M}\]
                    \State \textbullet $E_b^{t}$ remains empty 
                    \EndIf
              \EndIf
            \State \textbullet return $\tilde{r_t}$,$E_b^{t}$
\end{algorithmic}
\end{algorithm}
where \code{M} is a penalty to ensure that satisfying all constraints are not penalizing the reward. 
\section{Design of experiments}
\label{sec:Designofexperiments}
\subsection{Tools for performance comparisons}
\label{sec:toolsforperformance}

Measuring the performance of a multi-objective algorithm presents a more intricate challenge compared to evaluating a single-objective one. This complexity arises because a single scalar value cannot provide a comprehensive quantitative assessment of the algorithm's quality. Because of this complexity, it is always better to utilize multiple ones to rigorously assess the performances of multi-objective approaches \cite{ishibuchi2016sensitivity}. In the literature, the four most utilized are the HV, the $\epsilon$-indicator ($I_{\epsilon}^+$), the Generational Distance (GD), and the Inverted Generational Distance (IGD) \cite{riquelme2015performance}. The HV is a unary metric that only requires a reference point that should be dominated by all points in $\mathcal{P}^{\star}$, which is the nadir in \cite{mawdsley2022incore}. A higher HV is better. The other three are binary metrics that require the knowledge of all the elements of the approximate Pareto-optimal set  (although some alternatives exist \cite{riquelme2015performance}). For the latter three a lower value is better. Similarly, because the IGD is not Pareto-compliant, i.e., some set B dominated by another set A can have a better IGD value (i.e., lower value) than A \cite{ishibuchi2016sensitivity}, we have also added the IGD$^{+}$. The IGD$^{+}$ and provide better assessment than its IGD counterpart because it depends less on the distribution of the reference solution set \cite{ishibuchi2016sensitivity} (i.e., it is weakly Pareto-compliant). In accordance with existing literature, we utilize the Pareto front generated by running all algorithms as the reference set of solutions \cite{mawdsley2022incore,wang2022investigating}. We will also consider the C\_metric(A,Z) and $I_C$, which is the average number of point in set A dominated by points in the reference set Z and the number of them, respectively. A higher value the better. Furthermore, it's important to note that not all Pareto-optimal points are of equal importance \cite{kumar2016multiobjective}. Even if we have a set of Pareto-optimal solutions, we could propose a scheme to offer the best solution for the decision-maker. We will use the entropy method as described in \cite{kumar2016multiobjective}.

In the classical benchmark study presented in Section \ref{sec:testingapproaches}, we provide the HV and cardinality of Pareto-optimal solutions (via the $I_C$). However, we do not provide the $I_{\epsilon}^+$, neither GD, IGD, IGD$^{+}$, nor C\_metric in this context because the Pareto front is known, generated automatically, and depends on the desired number of points. The primary objective is to determine whether the points generated by our algorithm are dominated by these known solutions. On the other hand, for the PWR benchmark discussed in Section \ref{sec:applicationofPEARLtopwr}, we will utilize all the mentioned metrics to comprehensively assess algorithm performance.

\subsection{Test sets}
\label{sec:testsets}
\subsubsection{Classical benchmarks}
\label{sec:classicallbenchmarks}
To demonstrate the validity and performance of the developed algorithms, we will subject them to testing across various multi-objective benchmarks, both in unconstrained and constrained settings. For the unconstrained benchmarks, our selection includes the dtlzX test suite \cite{deb2002scalable}, specifically dtlz1, dtlz2, dtlz3, dtlz4, dtlz5, dtlz6, and dtlz7. dtlz1 is an M-objective problem with linear Pareto-optimal front. dtlz2 presents a dense Pareto front, while dtlz3 assesses an algorithm's ability to converge to a global pareto front with many local optima an MOO approach can converge to. dtlz4 assesses an algorithm's ability to maintain a well-distributed set of solutions. dtlz5 exhibits an optimal front in the form of a curve, and dtlz6 introduces added complexity with a non-linear distance function constraint g, making convergence against the Pareto-optimal curve more challenging. dtlz7 features disconnected Pareto-optimal regions, thereby evaluating the algorithm's capability to maintain sub-populations across different Pareto-optimal regions \cite{deb2002scalable}. 

In terms of constrained benchmarks, we opted to utilize classical benchmarks namely the cxdtlzy \cite{jain2014anevolutionaryII} and ctpx \cite{deb2001constrained} suites, which detailed can be found on pymoo's website \cite{blank2020pymoo}. To vary the shape and type of constrained spaces, we chose to study c1dtlz1 and c1dtlz3 (type 1 in \cite{jain2014anevolutionaryII}), c2dtlz2 (type 2 in \cite{jain2014anevolutionaryII}), and c3dtlz4 (type 3 in \cite{jain2014anevolutionaryII}). The ctpx test suite \cite{deb2001constrained} is also interesting. In ctp1, the infeasible region is very near the Pareto front, ctp2 is an ensemble of discrete Pareto regions, ctp3 reduces the feasible Pareto region to a single point, while ctp4 increases the difficulty of ctp3 by making the transition from continuous to discontinuous feasible spaces further away from the Pareto-optimal points \cite{deb2001constrained}.

Fortunately for these cases, we have access to the optimal Pareto front. Therefore, to assess the performance of the PEARL algorithms, we will utilize the Pareto front available in pymoo \cite{blank2020pymoo}. Following the taxonomy of pymoo, the dimension of the input space is $n_x = n_{obj} + k - 1$, where $n_{obj}$ is the number of objective for a particular problem and k is arbitrary. Because they are hard problems to solve, we choose k equal to 1 for dtlz1, dtlz3, c1dtlz1, and c1dtlz3. For the other unconstrained and constrained test cases we have k equal to 10 and 5, respectively.

\subsubsection{PWR multi-objective problems}
\label{sec:pwrmultiobjectiveproblems}
While the body of literature on this topic may not be as extensive as that for single-objective optimization, it has been a prevalent practice in PWR research to address multifaceted optimization problems involving multiple objectives and constraints. For instance in \cite{parks1996multiobjective}, the authors solved a three-objectives constrained PWR optimization problem with objectives such as minimizing the feed enrichment, maximizing the End-of-Cycle (EOC) burnup, and minimizing the radial form factor. They had a constraint on the form factor but did not explicitly solved for it. Similarly in \cite{park2014multicycle}, the authors solved a two-objective constrained problem with objectives to maximize $L_C$ and minimize the radial peaking factor $F_{xy}$ or EOC burnup (of FAs that will be loaded into the last cycle) for economic considerations, while satisfying constraints: $F_q$, pin discharge burnup, hot zero- and full-power MTC, and constraints on the two objectives. \cite{kropaczek2019largescale} leveraged a penalty-free constrained annealing method to solve a Two-objective core LP problem with objectives to maximize the cycle energy and minimize the vessel fluence, while satisfying constraints.  \cite{schlunz2016acomparative} derived a set of 16 multi-objective constrained problems for the in-core fuel management of the SAFARI-I research reactor. They have 7 types of objectives they combined within three classes.  However, being a research reactor, 5 out of 7 (e.g., Mo99 production) were not of interests for us. They however considered maximizing the $L_C$ while minimizing the relative core power peaking factor. Constraints on shutdown margin and peaking factor were also considered. Their goal was simply to compare the performance of the algorithms rather than gaining managerial insights on the problem at hand. Nevertheless, they solved problems up to four objectives, which is less common in literature but closer to real-world application.  Lastly, \cite{mawdsley2022incore} studied four different unconstrained optimization problems based on three different objectives, such as maximizing the EOC $C_b$ versus minimizing the pin power peaking $F_q$ and maximizing discharge burn-up.

These objectives were studied to trade-off economy (e.g., minimize feed enrichment, maximize $L_C$...) with safety constraints (e.g., minimize the peaking factors). By contrast, for the problem at hand \cite{seurin2022pwr,seurin2024assessment}, we utilized the LCOE as the main objective that encapsulates trade-offs between feed enrichment, $L_C$, and average core burnup, while the constraints added a penalty to the objective. Nonetheless, the MOO approach provides access to the granularity of the trade-offs, which can enhance decision-making. Here therefore, we replace the LCOE as the main objective by the $F_{\Delta h}$, the cycle length $L_C$, and the average feed enrichment in the core. The first parameter serves not only as a safety parameter that controls the margin to Departure from Nucleate Boiling (DNB) but can also be seen as a factor which reduces the thermal stresses in the core, limits fuel failure, and therefore increase the potential residence time of the FAs. The other two are operational and economical Figure Of Merits (FOMs). For the constrained case the constraints are the same as in \cite{seurin2022pwr,seurin2024assessment}, namely $F_q$, $C_b$, and peak pin burnup $Bu_{max}$. These problems will be studied in Section \ref{sec:unconstrainedbenchmarkproblems} and in Section \ref{sec:constrainedbenchmarkproblems} for the unconstrained and constrained PEARL, respectively.

\subsubsection{Legacy Approaches used for comparison}
\label{sec:legacyapproach}

For the core loading pattern optimization problem, the most widely leveraged heuristic-based algorithms are Genetic Algorithm (GA), Simulated Annealing (SA), and Tabu Search (TS) \cite{kropaczek2019largescale,seurin2023can,seurin2024surpassing} whether in single- and multi-objective settings. Therefore,  to compare every version of PEARL (i.e., PEARL-e, PEARL-\code{$\epsilon$}, and PEARL-NdS) against legacy approaches for the unconstrained case and C-PEARL-NdS with the formulation in equation \ref{eq:NucCoreObj} for the $CV$ in the constrained case, we adopted their multi-objective extensions: 

\begin{enumerate}
    \item 
    For GA we utilize NSGA-II \cite{deb2002afast} (which constrained version was not developed) and NSGA-III \cite{jain2014anevolutionaryI,jain2014anevolutionaryII}. They are often considered the state-of-the-art GA-based approach for multi-objective optimization \cite{stewart2021asurvey} and are heavily leveraged in the field of nuclear science and engineering \cite{zeng2020development,pevey2021multiobjective}. 
    \item 
    For SA, we use the formulation of \cite{park2009multiobjective}, where the acceptance of the new candidate hinge upon Pareto equivalence relationships between this solution and a buffer of stored solutions, as well as the $CV$. In case the previous $x$ and new candidates $\mathcal{N}(x)$ are both infeasible, the probability of acceptance is $\rho \sim e^{\frac{CV(x) - CV(\mathcal{N}(x))}{T}}$. In \cite{park2009multiobjective}, the Temperate T is constant, but we adopted an adaptive formulation that changes T depending on the values of each constraint and objective as in other core loading pattern optimization works \cite{kropaczek2019largescale,kropaczek2009copernicus}. 
    
    \item For TS, we draw inspiration from recent work on unconstrained MOO with TS \cite{mawdsley2022incore}, which we extend to constrained optimization with \cite{jaeggi2005amultiobjective}; The Ph.D. work in \cite{mawdsley2022incore} is built on top of \cite{jaeggi2005amultiobjective} and extended successfully the single-objective application of TS to core loading pattern optimization from \cite{hill2015pressurized} within the PANTHER framework. The advantage was to allow unrestricted application of any kind of physic code and algorithm to their core loading problem rather than limiting it to the use of Generalized Perturbation Theory  (GPT) \cite{kropaczek1991incore}. Nevertheless, this work did not consider parallel implementation, nor did it implement constraints. The parallel implementation is explicated in a concurrent paper \cite{seurin2024surpassing} implementing parallel Tabu Search (TS) for NEORL \cite{radaideh2023NEORL}. Moreover, our constraints' handling approach is adopted from \cite{jaeggi2005amultiobjective}. There, they assumed that all non-feasible solutions are Tabu, meaning that solutions are discarded until a feasible solution is found. We anticipate that we would have difficulties finding a feasible solution for real-world problems by random sampling. Indeed, our experience with single-objective and unconstrained multi-objective demonstrated that downhill moves are favorable for future gains. Therefore, we implemented two options: One is the same as discussed above. For the other one, we utilize the $CV$ in a constrained non-domination criterion as described for NSGA-III and SA. In practice, we found that the second option may work better for the some continuous problems studied. Lastly, it is common for TS algorithms to have a Medium Term Memory (MTM) \cite{mawdsley2022incore}. A MTM stores a certain number of  high-quality moves that were seen but not taken (to the benefit of an even better one). We control this by the hyper-parameter \code{intensification}, which represents the probability of selecting a move from the MTM or the best current move. 
\end{enumerate}

 To sum up, the multi-objective versions of SA and TS are similar to the single-objective ones, where the magnitude of the objective function is replaced by a comparison with a modified non-domination criterion as described in algorithm \ref{alg:modifiednondomination}. This criterion is used also in NSGA but for selecting the parent population \cite{jain2014anevolutionaryII}:
 
\begin{algorithm}[H]
  \small
    \caption{Pseudocode for modified non-domination criterion for constrained multi-objective optimization}
    \label{alg:modifiednondomination}
    \begin{algorithmic}[1]
    \State \textbf{Input}: E\_prev, E
    \State \textbf{Ouput}: E\_prev $\succ$ E
     \If{CV(E) $>$ 0 and CV(E\_prev) == 0}
     \State \textbullet return True
     \EndIf
    \If {CV(E) == 0 and CV(E\_prev) $>$ 0}
        \State \textbullet Return False
    \EndIf
    \If{CV(E) $>$ 0 and CV(E\_prev) $>$ 0}
        \If{SA}
        \State \textbullet Compute $\rho \sim e^{\frac{CV(E\_prev) - CV(E)}{T}}$
        \State \textbullet Compute $\epsilon \sim \mathcal{N}(0,1)$
        return $ \epsilon \le \rho$
        \Else
            \If{CV(E) $>$ CV(E\_prev)}
            \State \textbullet Return True
            \Else
            \State \textbullet Return False
            \EndIf
        \EndIf
    \EndIf
    \If{CV(E) == 0 and CV(E\_prev) == 0}
     \State  \textbullet use the classical unconstrained domination criteria
     \EndIf
\end{algorithmic}
\end{algorithm}

Additionally,  parallelism is implemented at the evaluation of each candidate solution in NSGA-II and NSGA-III. For both SA and TS, one processor is assigned to each Markov Chain \cite{seurin2024surpassing}. There, each chain carries a local buffer corresponding to some high-quality solutions that new solutions generated must be compared with as in PEARL (see Section \ref{sec:unconstrainedPEARL}) or multi-objective SA \cite{park2009multiobjective} and PESA \cite{konak2006multiobjective}.  Nevertheless, we cannot restart each chain with one solution based on the magnitude of the best solutions found as in single-objective \cite{seurin2024surpassing} but we must account for all solutions in the buffer. Therefore, at the end of the chain, the global Pareto front across all of them is first computed and a random sample is drawn to restart each.

We also defined PEARL-NdS (\code{crowding2}) and PEARL-NdS (\code{niching2}) to compare the performance of a pure ranking versus a smoother satisfaction of the constraints with $CV$ as a penalty in PEARL-NdS (\code{crowding}) and PEARL-NdS (\code{niching}). All of the algorithms described in this section as well as the variants of PEARL will be available in a further release of NEORL \cite{radaideh2023NEORL}.

\section{Results and analysis}
\label{sec:results}

\subsection{Testing the approaches on test suites}
\label{sec:testingapproaches}
\subsubsection{Unconstrained test suites}
\label{sec:unconstrainedtestsuites}
As described in Section \ref{sec:classicallbenchmarks}, we start by evaluating the performance of the derived algorithms on seven classical benchmarks namely dtlz1, dtlz2, dtlz3, dtlz4, dtlz5, dtzl6, and dtlz7. The envelope-based approach PEARL-e described in algorithm \ref{alg:PEARLepeusocode} is characterized by the generation of different rays every \code{n\_steps}. Therefore, we must illustrate the role of the rays and the shape of the Dirichlet distribution (by varying values of \code{$\alpha$}) on the optimal set of solutions found as well. We also conducted performance comparisons with four legacy approaches, specifically NSGA-II, NSGA-III, SA, and TS. It is worth noting that the computational overhead associated with training and subsequently applying the neural network during inference is likely to be overshadowed by the computational demands of the physics code used in real-world systems, as we observed in our single-objective PWR application \cite{seurin2023can}. Consequently, we have not addressed the latter aspect in this paper.

In-house experiments led to chose a value of 256 for $\code{n\_steps}\times \code{ncores}$ for PPO and PEARL, which amount to \code{n\_steps} equal to 32 for \code{ncores} equal to 8. We also use \code{n\_steps} equal 32 for the PEARL-e in the generation of rays and  \code{$\nu$} equal to 0.05 for PEARL-\code{$\epsilon$}. We compute optimizations up to 10,000 steps for all but dtlz1, dtlz3, dtlz6, and dtlz7 for which we compute 20,000 steps. \code{normalized\_obj} is set to False. We use cosine similarity \code{cos} for the \code{uniformity} and \code{$\lambda$} = 1 for the five PEARL-e cases but one for which we use $\lambda = 0$ (first five rows in Tables \ref{tab:unconstrainedbenchmarkshvdtlz1dtlz4}, \ref{tab:unconstrainedbenchmarkshvdtlz5dtlz7}, \ref{tab:unconstrainedbenchmarksicdtlz1dtlz4}, and \ref{tab:unconstrainedbenchmarksicdtlz5dtlz7}). The input space $\mathcal{A}$ is such that $\mathcal{A} = [0,1]^{n_x} \in \mathbb{R}$. The reference point for the calculation of the HVs is arbitrarily chosen to be $[3,3,3]$ for all but dtlz7 for which we choose $[3,3,7]$. For statistical rigour, we provide the average and standard deviation $\sigma$ over 20 experiments for each. The results are given in Tables \ref{tab:unconstrainedbenchmarkshvdtlz1dtlz4} and \ref{tab:unconstrainedbenchmarkshvdtlz5dtlz7} for the HV and Tables \ref{tab:unconstrainedbenchmarksicdtlz1dtlz4} and \ref{tab:unconstrainedbenchmarksicdtlz5dtlz7} for the $I_C$.

\begin{table}[H]
    \centering
    \caption{Different HV and associated standard deviation $\sigma$ over 20 experiments using different versions of PEARL against SO-based legacy approaches. For PEARL-e we show different combinations of \code{$\alpha$}. The best cases are highlighted in red. Friedman statistical test is significant with p $<<$ 0.05.}
    \begin{tabular}{c|c|c|c|c}
\hline
Algorithms  & dtlz1  & dtlz2	& dtlz3   &     dtlz4\\
             \hline
\makecell{PEARL-e \\$\lambda = 0$} & 26.72/	0.14 & 25.96/0.16 &	23.53/	3.04&25.81/0.14\\
\makecell{PEARL-e \\$\alpha = [0.5,0.5,0.5]$} & 26.82/	0.11 & 26.36/0.03	 & 24.68/	1.44 &26.16/0.08 \\
\makecell{PEARL-e \\ $\alpha = [1,1,1]^{\star}$} & 26.71/	0.17 & \textcolor{red}{26.40/0.01}	 & 23.15/	1.93 & 26.24/0.06 \\
\makecell{PEARL-e \\ $\alpha = [10,10,10]$} &  26.49/	0.41 & 26.38/0.01	& 21.35/	5.37 & 26.31/0.05\\
\makecell{PEARL-e \\ $\alpha = [0.99,2.0,2.0]$} & 	26.73 / 0.12 & \textcolor{red}{26.40/0.01} &23.90	/ 2.28
	 & 26.26/0.08 \\
PEARL-\code{$\epsilon$} & 26.93/	0.02  &  26.34/0.04  &  26.00/	0.14 &	26.16/0.05\\
\makecell{PEARL-NdS \\ (\code{niching}) } & 26.93/	0.01 & 26.33	/ 0.01 & 25.98/	0.16 &26.20	/ 0.03 \\
\makecell{PEARL-NdS \\ (\code{crowding}) } & 26.93/	0.02 & 26.36/0.04	 & 26.02/	0.14 &26.15/0.07  \\
NSGA-II	  & \textcolor{red}{26.96/	0.04} &  26.340/0.01&26.36/ 	0.01	 &26.35/	0.01 \\
NSGA-III	 & \textcolor{red}{26.96/	0.04} &  26.37/0.01 & \textcolor{red}{26.37/	0.01}	 &\textcolor{red}{26.36/0.01} \\
SA & 26.69/	0.27 & 26.04	/ 0.07 &23.07/	4.39 & 25.07 / 0.65 \\
TS & 26.88/	0.06 & 26.03	/ 0.09 & 25.67/	0.65 & 23.82 /	0.98\\
\end{tabular}
    \label{tab:unconstrainedbenchmarkshvdtlz1dtlz4}
\end{table}
      \begin{tablenotes}
      \item  \scriptsize$^{\star}$this case corresponds to a uniform distribution.
    \end{tablenotes}
\begin{table}[H]
    \centering
    \caption{Different HV and associated standard deviation $\sigma$ over 20 experiments using different versions of PEARL against SO-based legacy approaches. For PEARL-e we show different combinations of \code{$\alpha$}. The best cases are highlighted in red. Friedman statistical test is significant with p $<<$ 0.05.}
    \begin{tabular}{c|c|c|c}
\hline
Algorithms  &     dtlz5	    &    dtlz6	    &    dtlz7\\
             \hline
\makecell{PEARL-e \\$\lambda = 0$} & 23.31/0.22	&21.86/4.31	&\textcolor{red}{33.47/0.11} \\
\makecell{PEARL-e \\$\alpha = [0.5,0.5,0.5]$} & 23.86/0.03	&\textcolor{red}{24.00/0.001}	&33.27/0.21 \\
\makecell{PEARL-e \\ $\alpha = [1,1,1]$} & 23.91/0.02	&23.99/0.02	&33.34/0.32 \\
\makecell{PEARL-e \\ $\alpha = [10,10,10]$} &  23.87/0.02	&23.99/0.001	&33.17/0.20\\
\makecell{PEARL-e \\ $\alpha = [0.99,2.0,2.0]$} &  23.92/0.01 &	23.97/0.12 &	33.09/0.26 \\
PEARL-\code{$\epsilon$} & 23.86/0.04&	23.76/0.50&	33.35/0.18 \\
\makecell{PEARL-NdS \\ (\code{niching}) } & 23.79/0.03 & 23.99/0.00 & 33.37 / 0.08 \\
\makecell{PEARL-NdS \\ (\code{crowding}) } & 23.86/0.04	&\textcolor{red}{24.00/0.01}	&33.15/0.32    \\
NSGA-II	  & \textcolor{red}{23.96/0.004}&	23.97/0.003&	32.31/1.75 \\
NSGA-III	 & 23.93/0.02&	23.93/0.01&	32.58/0.19 \\
SA &  23.73	\ 0.06 & -289.85	/ 12.09 & 20.59 /	1.82 \\
TS &  23.74 \ 0.16 & -92.02 / 34.46 & 30.17 /	0.55\\
\end{tabular}
    \label{tab:unconstrainedbenchmarkshvdtlz5dtlz7}
\end{table}


\begin{table}[H]
    \centering
    \caption{Different $I_C$ and associated standard deviation $\sigma$ over 20 experiments using different versions of PEARL against SO-based legacy approaches. For PEARL-e we show different combinations of \code{$\alpha$}. The best cases are highlighted in red. Friedman statistical test is significant with p $<<$ 0.05.}
    \begin{tabular}{c|c|c|c|c}
    \hline
Algorithms&  dtlz1	 &  dtlz2&  dtlz3	 	      &  dtlz4\\
\hline
\makecell{PEARL-e \\ $\lambda = 0$} &	1.4/	1.2	 &   0.05/0.0	&  0.9	/ 0.94  & 0.0	/ 0.0 \\
\makecell{PEARL-e \\ $\alpha = [0.5,0.5,0.5]$}& 1.2	/0.93 &	542.65/265.60 & 1.25 /	0.99 &	2.2/2.99\\
\makecell{PEARL-e \\ $\alpha = [1,1,1]$}&	0.8	/0.75 &1478.5/253.20 &0.4	/0.9&	11.55/8.26 \\
\makecell{PEARL-e \\ $\alpha = [10,10,10]$}&0.85/	0.79	 &\textcolor{red}{3359.0/828.08} &0.6 / 0.8&	32.7/16.50\\
\makecell{PEARL-e \\ $\alpha = [0.99,2.0,2.0]$} &1.3/1.38	 &1908.8/279.30 &0.7/ 0.78	& 14.1/9.22 \\
PEARL-\code{$\epsilon$}	  & 3.95 / 1.72  & 478.45/238.1 &2.05/	1.40&2	1.5/1.12 \\
\makecell{PEARL-NdS \\ (\code{niching})} & 4.6 / 2.27 & 26.55/ 9.02 &  2.35 / 1.28 & 2.05/2.13 \\
\makecell{PEARL-NdS \\ (\code{crowding})}  & 5.7	/3.03 &503.35/246.28 &	2.35/	1.8&2.05/2.52\\
NSGA-II	  &  \textcolor{red}{51.8/	11.91} &46.65/3.86 & \textcolor{red}{53.25/	0.8}& 	\textcolor{red}{46.3/1.85}\\
NSGA-III	  & 49.9/	11.71  &43.6/2.33 &52	/2.81	& 34.3/3.82\\
SA & 36.5/	51.89& 11.95 /	6.52  & 24.4/	48.11 & 14.6 / 3.64\\
TS & 79.1/	53.63 & 47.65 /	13.39 & 45.4/	53.83 & 5.35 /	2.15\\
\\
    \end{tabular}
    \label{tab:unconstrainedbenchmarksicdtlz1dtlz4}
\end{table}

\begin{table}[H]
    \centering
    \caption{Different $I_C$ and associated standard deviation $\sigma$ over 20 experiments using different versions of PEARL against SO-based legacy approaches. For PEARL-e we show different combinations of \code{$\alpha$}. The best cases are highlighted in red. Friedman statistical test is significant with p $<<$ 0.05.}
    \begin{tabular}{c|c|c|c}
    \hline
Algorithms&     dtlz5	   &     dtlz6	   &     dtlz7\\
\hline
\makecell{PEARL-e \\ $\lambda = 0$} &     33.75/7.25& 1851.5/1780.1 & \textcolor{red}{5217.8/1122.9} \\
\makecell{PEARL-e \\ $\alpha = [0.5,0.5,0.5]$}& 35.1/4.59	 &  12043.0/1273.8	& 4434.8/1593.7 \\
\makecell{PEARL-e \\ $\alpha = [1,1,1]$}& 38.05/8.35	 &   13602/1807.0	& 4240.6/1223.7 \\
\makecell{PEARL-e \\ $\alpha = [10,10,10]$}& \textcolor{red}{371.6 /219.57}	 & \textcolor{red}{14362.0/1027.1}	& 2293.7/1232.7 \\
\makecell{PEARL-e \\ $\alpha = [0.99,2.0,2.0]$} & 51.05/14.04	& 13375./1115.1 &	3697.4/1262.1 \\
PEARL-\code{$\epsilon$}	  & 35.25/4.58	 & 10938./3643.5	& 3976.9/1187.9 \\
\makecell{PEARL-NdS \\ (\code{niching})} &  28.35/6.01 &  499.1/1.3&352.85/13.87\\
\makecell{PEARL-NdS \\ (\code{crowding})}  &  34.8/5.80&  11468./2449.8	& 4145.7/1154.8 \\
NSGA-II	  &  48.1/3.97	 &  57.3/0.78  &  10.75/6.92 \\
NSGA-III	  & 43.55/3.89&	51.4/2.44	   &  16.25/5.85 \\
SA &  0.1	/ 0.44 & 0 / 0 & 0 / 0\\
TS & 19.15	/ 8.18 & 0 / 0 & 0/0\\
\\
    \end{tabular}
    \label{tab:unconstrainedbenchmarksicdtlz5dtlz7}
\end{table}

As showcased in Tables \ref{tab:unconstrainedbenchmarksicdtlz1dtlz4} and \ref{tab:unconstrainedbenchmarksicdtlz5dtlz7} all PEARL algorithms are capable of recovering points on the front. Additionally, we see with Tables \ref{tab:unconstrainedbenchmarkshvdtlz1dtlz4} and \ref{tab:unconstrainedbenchmarkshvdtlz5dtlz7} that the PEARL-e method seems to perform better, exacerbating large HV (and $I_C$ as seen in Tables \ref{tab:unconstrainedbenchmarksicdtlz1dtlz4} and \ref{tab:unconstrainedbenchmarksicdtlz5dtlz7}) for different \code{$\alpha$} distributions. Moreover, the HV are quite close in practice to the best HV found in all problems for all versions of PEARL except for PEARL-\code{$\epsilon$}. The indicator utilized for the latter is the basic $I_{\epsilon}$, which suffers from the \textit{edge effect} \cite{yang2019amanyobjective}, which states that an indicator-based method privileges individuals around the boundary of the front. As a result, the solution found by PEARL-\code{$\epsilon$} could be localized and the resulting front less dispersed compared to its PEARL-e and PEARL-NdS counterpart. We have chosen not to pursue the optimization of this specific aspect, but we welcome any interested scientists to explore and enhance it further. For the other versions, our findings affirm the versatility of PEARL in addressing various problems with different front distributions. The significant disparities in cardinality $I_C$ compared to classical approaches arise from the buffer allocated to each agent, which is not shared in our implementation (i.e., we can have up to \code{ncores}x\code{$\kappa$} solutions, equal to 512 for PEARL-NdS and PEARL-$\epsilon$ but no limit for PEARL-e). We also observe that the best algorithm depends on the problem, or the response surface, which is an illustration of the no-free lunch theorem for optimization \cite{wolpert1997no}.


Additionally, while PEARL-e proves to be an efficient method (high HV on 6 out of 7 unconstrained benchmarks), its performance depends on the choice of \code{$\alpha$}, hence we must assume some knowledge of the location of the Pareto front. This accentuates the challenge of potentially introducing computational overhead to adapt the rays to the specific problem at hand. This complexity may arise, particularly when the Pareto front is unknown. However, when a designer possesses some knowledge of the Pareto surface, fine-tuning the Dirichlet distribution with \code{$\alpha$} could yield significant advantages. Conversely, the PEARL-NdS methods systematically found Pareto-optimal solutions without the need for any assumptions regarding ray distributions tailored to a specific front. This flexibility is highly beneficial for real-world practitioners who seek to utilize these tools without delving into problem-specific tuning. 

Lastly, we can analyze the NTs results for the HVs in Tables \ref{tab:unconstrainednemenyidtlz1dtlz4} and \ref{tab:unconstrainednemenyidtlz5dtlz7}. For the sake of brevity, we've only included the columns corresponding to the algorithm with the highest average HV across the 20 experiments. For each case where PEARL is superior (dtlz2, dtlz6 and dtlz7), the NTs are successful against SO-based legacy MOO methods. On the other hand, for problems for which NSGA-based approaches are superior (dtlz1, dtlz3, dtlz4, and dtlz5), some NTs failed against PEARL (e.g., PEARL-e with \code{$\alpha$} $= [0.99,2.0,2.0]$ for dtlz5). This implies that there are no statistically significant differences between these algorithms on the latter problems but for the former problems PEARL is statistically better. This further underscores the potential for PEARL to serve as a viable alternative to legacy approaches in unconstrained MOO scenarios.

\begin{table}[H]
    \centering
     \caption{NT for the highest HVs cases: unconstrained tests suite. In red, the failed statistical tests ($\alpha > 0.1$).}
    \begin{tabular}{c|c|c|c|c}
    \hline
	& \makecell{dtlz1 \\ NSGA-II} & \makecell{dtlz2 \\PEARL-e \\ $\alpha = [1,1,1]$}&	\makecell{dtlz3 \\ NSGA-III} & \makecell{dtlz4\\NSGA-III} \\
 \hline
PEARL-e $\lambda = 0$           &0.001& 0.001&0.001	&0.001\\
PEARL-e $\alpha = [0.5,0.5,0.5]$	&0.001&0.001&	0.001	&0.001\\
PEARL-e $\alpha = [1,1,1]$      &0.001& -	 &	0.001  & 0.001\\
PEARL-e $\alpha = [10,10,10]$	  &0.001 & \textcolor{red}{0.264}&	0.001 &\textcolor{red}{0.324}\\
PEARL-e $\alpha = [0.99,2.0,2.0]$&	0.001 &\textcolor{red}{0.9} &	0.001  & 0.005\\
PEARL-\code{$\epsilon$}	      & \textcolor{red}{0.9}& 0.001&\textcolor{red}{0.053}	&	0.001\\
PEARL-NdS (\code{niching})	  & \textcolor{red}{0.9}& 0.001& \textcolor{red}{0.16} &\textcolor{red}{0.054}		\\
PEARL-NdS (\code{crowding})	  & \textcolor{red}{0.68}& 0.002&\textcolor{red}{0.053}\\
NSGA-II	      &- & 0.001&	\textcolor{red}{0.9}	  &	 0.1\\
NSGA-III	  & \textcolor{red}{0.9}    & 0.002&-	&	-\\
SA &0.001&0.001& 0.001 & 0.001\\
TS &0.02& 0.001& 0.004 &0.003\\
    \end{tabular}
    \label{tab:unconstrainednemenyidtlz1dtlz4}
\end{table}

\begin{table}[H]
    \centering
     \caption{NT for the highest HVs cases: unconstrained tests suite. In red, the failed statistical tests ($\alpha > 0.1$).}
    \begin{tabular}{c|c|c|c}
    \hline
	& \makecell{dtlz5\\NSGA-II}&	\makecell{dtlz6\\PEARL-e \\ $\alpha = [0.5,0.5,0.5]$}&	\makecell{dtlz7\\ PEARL-e \\ $\lambda = 0$} \\
 \hline
PEARL-e $\lambda = 0$           &0.001	& -\\
PEARL-e $\alpha = [0.5,0.5,0.5]$	&0.001	&-	    &\textcolor{red}{0.212}\\
PEARL-e $\alpha = [1,1,1]$      &0.034	&\textcolor{red}{0.9}	  &  \textcolor{red}{0.819}\\
PEARL-e $\alpha = [10,10,10]$	  &0.001	&0.9	&0.04\\
PEARL-e $\alpha = [0.99,2.0,2.0]$&\textcolor{red}{0.357}	&\textcolor{red}{0.626}	&0.001\\
PEARL-\code{$\epsilon$}	      & 0.001	&\textcolor{red}{0.392}	&\textcolor{red}{0.691}\\
PEARL-NdS (\code{crowding})	  & 0.001	&\textcolor{red}{0.357}	&0.007\\
PEARL-NdS (\code{niching})	  &  0.001&	0.001	&	\textcolor{red}{0.9}\\
NSGA-II	      &-	    &0.001	&0.001\\
NSGA-III	  & \textcolor{red}{0.8192}&	0.001&	0.001\\
SA &0.001&0.001&0.001\\
TS &0.001& 0.001& 0.001\\
    \end{tabular}
    \label{tab:unconstrainednemenyidtlz5dtlz7}
\end{table}


\subsubsection{Constrained test suites}
\label{sec:constrainedtestsuites}

 We conduct similar experiments as in Section \ref{sec:unconstrainedtestsuites}, where we provide the HV and $I_C$. We are using 10,000 steps for all problems. The input space $\mathcal{A}$ is such that $\mathcal{A} = [0,1]^{n_x} \in \mathbb{R}$. The nadir is arbitrarily chosen to be $[3,3,3]$ and $[3,3]$ for the cxdtlzy and ctpx, respectively. \code{ncores} and \code{normalized\_obj} are set to 8 and False, respectively. We use \code{uniformity} as cosine similarity \code{cos} and $\lambda$ = 1. \code{$\alpha$} $= [10,10,10]$, \code{$\kappa$} = 64 for all but \code{niching2} for which we choose 1024, \code{M} equal to \code{$\kappa$}, and \code{$\nu$} = 0.05. We compare the approaches to the constrained version of NSGA-III, SA, and TS as described in Section \ref{sec:legacyapproach}. The results are given in Tables \ref{tab:clvsbasehvbenchmarkcdtlz}, \ref{tab:clvsbasehvbenchmarkctp}, \ref{tab:clvsbaseicbenchmarkcdtlz}, and \ref{tab:clvsbaseicbenchmarkctp}. For statistical rigour, we provide the average and standard deviation $\sigma$ over 20 experiments for each. 

Looking at the higher HV and $I_C$ for most \code{CL}-based PEARL, we see that utilizing the \code{CL} method (i.e., solving for the constraints first) shows great performance for all but the c2dtlz2 problem with the PEARL-e method. The hypothesis put forth in Section \ref{sec:CurriculumLearningandhierarchical}, suggesting to divide the problem into two simpler sub-problems, has indeed led to high performance for the RL agents. NSGA performs well throughout except for c1dtlz1 and c1dtlz3. NSGA-III \cite{jain2014anevolutionaryII} is also treating constraints first before dealing with the regular target FOMs. PEARL-NdS (\code{crowding}) and (\code{crowding}) present the highest HV for 4 out of the 8 benchmarks and the former the highest $I_C$ for 5 out of 8.  Furthermore, across all benchmark scenarios, the HV and $I_C$ are greater than SO-based techniques for at least one version of PEARL (with \code{CL}). It confirms therefore the potential of PEARL (with \code{CL}) to scale for different problems with different front distributions in the constrained case as well. 


\begin{table}[H]
    \centering
     \caption{Comparison of HV and associated standard deviation $\sigma$ over 20 experiments using different versions of 
     C-PEARL against SO-based legacy approaches. for the application to classical benchmarks. The best cases are highlighted in red. Friedman statistical test is significant with p $<<$ 0.05.}
    \begin{tabular}{c|c|c|c|c}
\hline
algorithm	 & c1dtlz1 & c1dtlz3 & c2dtlz2		&	c3dtlz4\\
\hline
PEARL-e		& 17.09/	9.88 & 18.48	7.76& 26.13	/ 0.03 			  &  23.23	/ 0.23 \\
PEARL-\code{$\epsilon$}		 &23.53/	5.50& 25.97	/ 0.28& 22.65	/ 2.56 		      &  23.09	/ 0.26 \\
\makecell{PEARL-NdS \\ (\code{niching})-\code{CL}}	 & 23.64/	5.51 & 25.84	0.26 &26.31	/0.02 & 23.19	/0.23 \\
\makecell{PEARL-NdS \\ (\code{crowding})-\code{CL}}	 & 24.45/	1.32 & \textcolor{red}{26.03	/ 0.18} &\textcolor{red}{26.30	/ 0.02} 			  &  23.10	/ 0.21\\
\makecell{PEARL-NdS \\ (\code{niching2})}	 & 24.21/	1.00 &17.99	0.05	& 26.15/0.02	& \textcolor{red}{23.56/0.21}\\
\makecell{PEARL-NdS \\ (\code{crowding2})} & \textcolor{red}{24.97/	0.88} & 17.97	\ 0.18 & 18.00	/0.03 & \textcolor{red}{23.56	/0.19} \\
\\
NSGA-III     & 3.37/	8.03 &   17.60	0.96 & 25.07 / 1.29                & 22.97	/ 1.83\\
SA & 17.80/	8.90 & 12.43	5.40 & 17.86 /	0.08 & 14.42 /	2.20\\
TS & 11.18/	11.19 & 15.23	3.51 & 18.00	/ 0.00 & 14.73 /	2.45\\
    \end{tabular}
    \label{tab:clvsbasehvbenchmarkcdtlz}
\end{table}

\begin{table}[H]
    \centering
     \caption{Comparison of HV and associated standard deviation $\sigma$ over 20 experiments using different versions of 
     C-PEARL against SO-based legacy approaches. for the application to classical benchmarks. The best cases are highlighted in red. Friedman statistical test is significant with p $<<$ 0.05.}
    \begin{tabular}{c|c|c|c|c}
\hline
algorithm	 &       ctp1			&            ctp2			&            ctp3			&            ctp4	\\
\hline
PEARL-e		& 	    7.20    / 0.01&	 7.71  / 	0.01	&7.61	/ 0.01&		7.04/	0.22 \\
PEARL-\code{$\epsilon$}		 &		6.21	/ 0.06	&	 \textcolor{red}{7.76  / 	0.004}	& \textcolor{red}{7.66	/ 0.01}	&	7.18	/ 0.23 \\
\makecell{PEARL-NdS \\ (\code{niching})-\code{CL}}	 &  7.21	/0.02 & \textcolor{red}{7.76	/0.00} & 7.65	/0.01 & 7.11	/0.21 \\
\makecell{PEARL-NdS \\ (\code{crowding})-\code{CL}}	 &		\textcolor{red}{7.22	/ 0.01} &	 7.75  / 	0.01	&7.64	/ 0.01	&	7.03	/ 0.18 \\
\makecell{PEARL-NdS \\ (\code{niching2})}	 & \textcolor{red}{7.22/0.01}	& 7.74/0.01	& 7.63/0.01	& \textcolor{red}{7.28/0.15} \\
\makecell{PEARL-NdS \\ (\code{crowding2})} & 7.22	/0.00 & \textcolor{red}{7.76	/0.00} & 7.67	/0.01 & 7.33	/0.18 \\
NSGA-III     &       6.96/0.05           & 7.73/0.01 &7.64 /0.02  & 6.63/0.53 \\
SA & 7.16 /	0.02 & 7.73 /	0.01  & 7.58  /	0.07  & 7.19 /	0.12\\
TS & 7.19 /	0.01 & \textcolor{red}{7.76/	0.01} & 7.63 /	0.04 & \textcolor{red}{7.28 /	0.07}\\
    \end{tabular}
    \label{tab:clvsbasehvbenchmarkctp}
\end{table}

\begin{table}[H]
    \centering
    \caption{Comparison of $I_C$ and associated standard deviation $\sigma$ over 20 experiments using different versions of 
    C-PEARL against SO-based legacy approaches for the application to classical benchmarks. The best cases are highlighted in red. All FTs succeeded with p = 0.00.}
    \begin{tabular}{c|c|c|c|c}
\hline
algorithm	 &  c1dtlz1 & c1dtlz3 & c2dtlz2		&	c3dtlz4\\
\hline
PEARL-e	 & 0.45/	0.80 & 0.35/	0.65&  10.6	/ 7.85    &   12.25 /	4.59 \\
PEARL-\code{$\epsilon$}		 & 2.15/	1.96& \textcolor{red}{3.70	/3.07}& 39.75	/ 49.46   &   8	  /  2.63\\
\makecell{PEARL-NdS \\ (\code{niching})-\code{CL}} & 2.25/	1.22 & 2.40	/1.46&184.65	/45.48 & 11.85/	3.07 \\

\makecell{PEARL-NdS \\ (\code{crowding})-\code{CL}	} & 2.55/	1.99 & 2.95/	1.69 &  \textcolor{red}{192.1	/ 59.84}   &   12.05	/4.33 \\
\makecell{PEARL-NdS \\ (\code{niching2})}	 & 2.70/	1.79 &0.00/	0&12.0/4.40	& 19.4/4.57\\

\makecell{PEARL-NdS \\ (\code{crowding2})} & \textcolor{red}{4.25/	1.44} & 0.00/	0 & 0	/0 & 17.15/	3.90 \\
NSGA-III     &0.00/	0.00 & 0.00/	0 & 34.15 / 7.21   & \textcolor{red}{37.20 / 13.97} \\
SA & 0.45/	1.75& 0.20	/ 0.68& 0 / 0  & 10	/ 3.11\\
TS & 0.00/	0.00 & 0.00/	0 &  0.8 / 2.66 & 10.45/	5.65\\
  \end{tabular}
    \label{tab:clvsbaseicbenchmarkcdtlz}
\end{table}

\begin{table}[H]
    \centering
    \caption{Comparison of $I_C$ and associated standard deviation $\sigma$ over 20 experiments using different versions of 
    C-PEARL against SO-based legacy approaches for the application to classical benchmarks. The best cases are highlighted in red. All FTs succeeded with p = 0.00.}
    \begin{tabular}{c|c|c|c|c}
\hline
algorithm	 &         ctp1			 &       ctp2			 &       ctp3		&	        ctp4		\\
\hline
PEARL-e	 & 45.45	/ 49.77	&	1.35/	0.48	&	1	 /   0		  &     \textcolor{red}{1/	0} \\
PEARL-\code{$\epsilon$}		 &		520	 /   0		&        9.8	  /  4.73	&	\textcolor{red}{1.05/	0.2}	&	\textcolor{red}{1	/0} \\
\makecell{PEARL-NdS \\ (\code{niching})-\code{CL}} & 344.8/	49.22 & 14.7	/4.86 & 1.05	/0.22 & 1.05/	0.22 \\

\makecell{PEARL-NdS \\ (\code{crowding})-\code{CL}	} & \textcolor{red}{321.7/	19.58}	& \textcolor{red}{18.15	/6.20} &	\textcolor{red}{1.05/	0.22}	&	\textcolor{red}{1/	0} \\
\makecell{PEARL-NdS \\ (\code{niching2})}	 & 83.35/12.86 &	2.6/1.53	& 1 /0.0	& \textcolor{red}{1 /0.0} \\

\makecell{PEARL-NdS \\ (\code{crowding2})} & 366.95/	13.99 & 13.4	/2.97 & 1	/0 & 1	/0 \\
NSGA-III     & 8.55 / 1.47 & 2.45 / 1.24 & \textcolor{red}{1.05}	/ 0.21 &0.95/ 0.22 \\
SA &  11.4 /	8.73 & 2.25	 / 1.813 & 0.00 / 0.00 & 0.05 /	0.22\\
TS & 26.75	/ 9.65 & 6.2 /	2.91 & 0.00 / 0.00 & 0.05 /	0.22\\
  \end{tabular}
    \label{tab:clvsbaseicbenchmarkctp}
\end{table}

Finally, we can analyze the NTs results for the HV in Tables \ref{tab:constrainednemenyicdtlz} and \ref{tab:constrainednemenyictp}. To streamline the presentation, we've focused on the columns corresponding to the algorithm with the highest average HV across the 20 experiments in each case. In every scenario, one version of PEARL consistently outperforms others by achieving the highest average HV.

Furthermore, in at least 6 out of the 8 benchmarks, the NTs demonstrate that PEARL surpasses the SO-based approaches. This underscores the potential for PEARL to serve as a successful alternative to legacy approaches in constrained MOO scenarios. Notably, the results are even more favorable in the constrained cases compared to the unconstrained ones. 

Thanks to the successful application of the different versions of PEARL on classical benchmarks, Section \ref{sec:applicationofPEARLtopwr} depicts their performance on a real-world PWR system.

\begin{table}[H]
    \centering
    \caption{NT tests for the highest HVs cases: constrained tests suite. In red, the failed statistical tests ($\alpha > 0.1$).}
    \begin{tabular}{c|c|c|c|c}
    \hline
& \makecell{c1dtlz1 \\ (\code{crowding2})}  &  \makecell{c1dtlz3 \\ (\code{crowding}) \\ with \code{CL}}  & \makecell{c2dtlz2 \\ (\code{crowding}) \\
with \code{CL}} &	\makecell{c3dtlz4 \\ (\code{niching2})}\\
 \hline
PEARL-e ($\alpha = [10,10,10]$) &0.001& 0.00&   0.00290	&\textcolor{red}{0.530}\\
PEARL-\code{$\epsilon$}	   &\textcolor{red}{0.9}& \textcolor{red}{0.9} &   0.001	&\textcolor{red}{0.129}\\
PEARL-NdS (\code{niching})-\code{CL}	   &\textcolor{red}{0.9}&  \textcolor{red}{0.9}&   \textcolor{red}{0.9}    &\textcolor{red}{0.264}\\
PEARL-NdS (\code{crowding})-\code{CL}	&\textcolor{red}{0.9}& - &   - &\textcolor{red}{0.11} \\
PEARL-NdS (\code{niching2})&\textcolor{red}{0.9}&	 0.001&   0.00357	&-\\
PEARL-NdS (\code{crowding2}) &-& 0.001 & 0.001 & \textcolor{red}{0.9} \\
NSGA-III	&0.001&  0.001   &   0.001	&\textcolor{red}{0.9} \\
SA &0.001&0.001& 0.001 & 0.001\\
TS &0.001&0.001& 0.001 & 0.001\\
    \end{tabular}
    \label{tab:constrainednemenyicdtlz}
\end{table}

\begin{table}[H]
    \centering
    \caption{NT tests for the highest HVs cases: constrained tests suite. In red, the failed statistical tests ($\alpha > 0.1$).}
    \begin{tabular}{c|c|c|c|c}
    \hline
&	\makecell{ctp1 \\(\code{crowding}) \\
with \code{CL}}&	\makecell{ctp2 \\ PEARL-\code{$\epsilon$}\\
with \code{CL}}&	\makecell{ctp3 \\ PEARL-\code{$\epsilon$}\\
with \code{CL}}&	\makecell{ctp4 \\ (\code{niching2})} \\
 \hline
 PEARL-e ($\alpha = [10,10,10]$)&	\textcolor{red}{0.147}&	0.003&	0.001&	0.039\\
PEARL-\code{$\epsilon$}	   &	0.001&	-	   & -	 &   \textcolor{red}{0.9}\\
PEARL-NdS (\code{niching})-\code{CL}	   &	\textcolor{red}{0.9}	 &   \textcolor{red}{0.9}&	\textcolor{red}{0.787}&	\textcolor{red}{0.147}\\
PEARL-NdS (\code{crowding})-\code{CL}	& -	 &   \textcolor{red}{0.9}&	\textcolor{red}{0.12}&	0.06\\
PEARL-NdS (\code{niching2})& \textcolor{red}{0.755}&	\textcolor{red}{0.357}&	0.020&	-\\
PEARL-NdS (\code{crowding2}) & \textcolor{red}{0.84} & \textcolor{red}{0.9} & \textcolor{red}{0.9} & \textcolor{red}{0.9}\\
NSGA-III	& 0.001&	0.046&	\textcolor{red}{0.188} &	0.001\\
SA & 0.001& 0.001& 0.001& \textcolor{red}{0.9}\\
TS & 0.001& \textcolor{red}{0.9} & 0.00 &\textcolor{red}{0.9}\\
    \end{tabular}
    \label{tab:constrainednemenyictp}
\end{table}
\subsection{Application of PEARL to PWR Optimization}
\label{sec:applicationofPEARLtopwr}
In our earlier research \cite{seurin2023pareto}, we observed that in unconstrained scenarios, both PEARL-NdS (\code{crowding}) and PEARL-NdS (\code{niching}) exhibited similar performance. Similarly, when using cosine similarity \code{cos} or KL-divergence \code{KL-div} in PEARL-e, the results were comparable. Consequently, for unconstrained cases, we will present only one version from each pair of methods. However, because of its superiority in the unconstrained setting, we will focus on PEARL-NdS in the context of constrained scenarios and provide all possible combinations of selection (i.e., PEARL-NdS with \code{CL}: \code{crowding},\code{niching},\code{crowding2},\code{niching2}).

\subsubsection{Unconstrained Benchmark problems}
\label{sec:unconstrainedbenchmarkproblems}

\begin{figure}[H]
    \centering
    \includegraphics[scale=0.55]{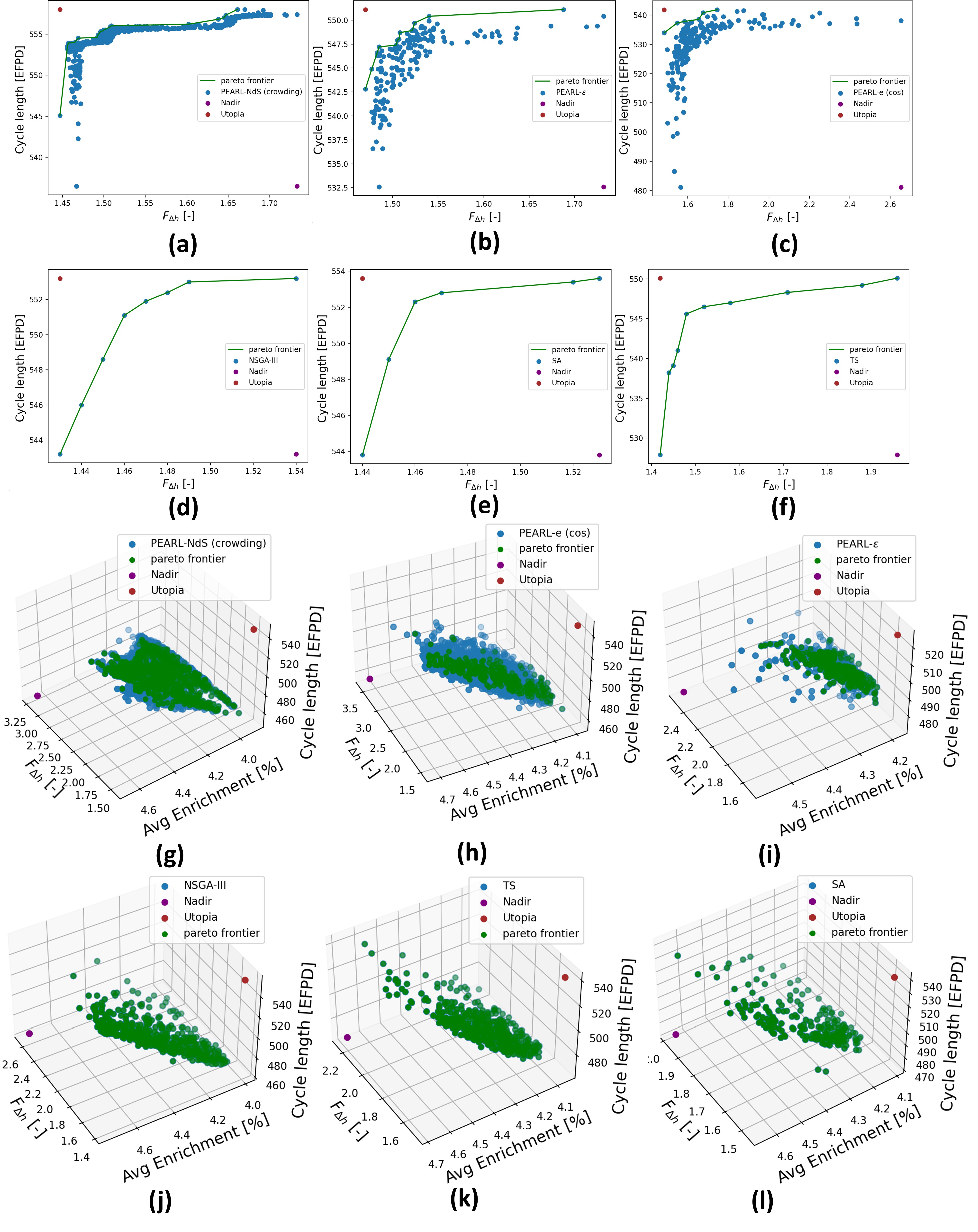}
    \caption{Two-obj ((a) to (f)) and Three-obj ((g) to (L))) solutions generated for all the algorithms utilized. The pareto front is a line for the Two-objective case, while it is drawn with scattered points for the Three-objective one.}
    \label{fig:bothuncsontrained}
\end{figure}

The details of the geometry, the decision space $\Gamma$, and the feasible space $\mathbb{O} \subset \Gamma$ are given in \cite{seurin2022pwr,seurin2024assessment}. As mentioned in Section \ref{sec:pwrmultiobjectiveproblems}, we decided to minimize the peaking $F_{\Delta h}$, maximize the $L_C$, and minimize the average enrichment in the core. The first test will utilize the first two objectives, while the second test will consider all three. It will be intriguing to observe how the envelope coverage changes as we are adding a third type of trade-off surface. There, we ran the RL agents (similar hyper-parameters for RL as in \cite{seurin2022pwr,seurin2024assessment}) from scratch up to 50,000 and 100,000 objective function evaluations for the Two- and Three-obj cases, respectively, while using 32 cores (i.e. 32 agents) on an Intel Xeon Gold 5220R clocked at 2.20 GHz. For cosine similarity as \code{uniformity} for PEARL-e, the option \code{normalized\_obj} is set to \code{False}. Similarly, we ran the NSGA-III on 32 cores, which results in 32 individuals in a population. For the GA portion of the NSGA-III algorithms, we used  the Evolutionary (\code{$\mu$},\code{$\lambda$}) Strategy (ES) \cite{beyer2002evolution}. We use \code{$\lambda$\_} equal to 32, \code{$\mu$} equal to \code{$\lambda$\_},\code{mutpb} to 0.3, and \code{cxpb} to 0.65. For SA and TS we use \code{nChain} equal to \code{ncores}. Specifically for SA we have \code{chain\_size} equal to 50,\code{lmbda} to 15, \code{alpha} to 2,\code{threshold} to 0.00, \code{chi} to 0.1, and
        \code{Tmin} to 0.005. Lastly for TS we have \code{penalization\_weight} equal to 6.0, \code{tabu\_tenure} to 162, \code{chain\_size} to 10, and \code{intensification} to 0.2. Since PEARL will propose more solutions from the different buffers, we will provide the Pareto front of all the solutions ever found from SO. Typically, we looked at all the solutions rather than the last population for NSGA or the last element of each chain for SA and TS (i.e., 32 of them for SO) for fairness. The individual results for the Two- and Three-obj cases are given in Figure \ref{fig:bothuncsontrained} (grouped in six figures each). 

\begin{figure}[H]
    \centering
    \includegraphics[scale=0.65]{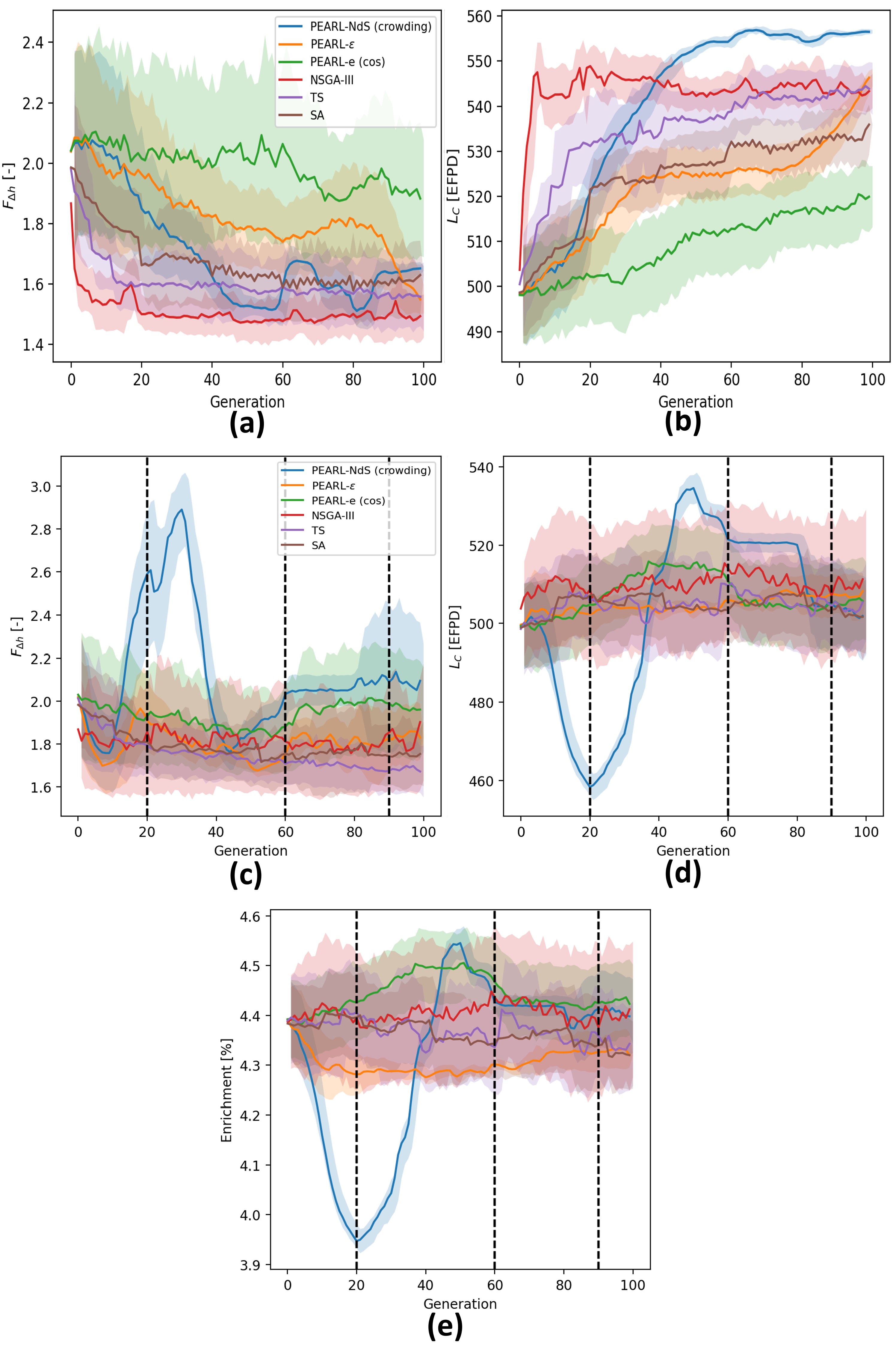}
    \caption{Evolution of the target FOMs for the Two-objective ((a) and (b)) and the Three-objective ((c) to (e)) problem over 100 generations for PEARL. We added vertical grid lines in the Three-objective case to help with visibility and interpretation of the results: (a) and (c) $F_{\Delta h}$, (b) and (d) $L_{cy}$, and (e) Average feed enrichment. The lighter areas cover the average $\pm \sigma$.}
    \label{fig:fomsunconstrained3d}
\end{figure}

In the Two-obj case, a trade-off between the $L_C$ and $F_{\Delta h}$ becomes apparent. There exists a trade-off for low values of the latter but it diminishes for high values. In the former case, the pattern may be asymmetric incurring high $F_{\Delta h}$ but also susceptible to suboptimal utilization of certain FAs, resulting in reduced cycle energy. The examples mentioned above provide straightforward first-order (i.e., linear) explanations. However, there are additional alternative solutions available by manipulating factors such as enrichment, burnable poison, and their placement within the core, which introduce numerous trade-offs at a more detailed level. Furthermore, when peaking factors are extremely low, the $L_C$ can undergo significant variations, even though it is feasible to achieve a comparable $L_C$ level for different values of $F_{\Delta h}$, ranging from approximately 1.55 to 2.2. This situation could potentially lead to substantial economic losses if the constraint on $F_{\Delta h}$ is overly stringent. The PEARL-e and PEARL-\code{$\epsilon$} approaches fail to find low $F_{\Delta h}$ and the latter preferentially generated higher $L_C$ (or lower $-L_C$), while the former is stuck in a low quality Pareto front. Figures \ref{fig:fomsunconstrained3d}.(a) and \ref{fig:fomsunconstrained3d}.(b) shows that all algorithms will increase the $L_{cy}$ but PEARL-e and PEARL-\code{$\epsilon$} stays with a relatively high $F_{\Delta h}$ compared to the others, while the SO-based approaches resulted in lower $F_{\Delta h}$ more rapidly than any other PEARL. These figures alone does not provide a clear indication of which approach (PEARL-NdS (\code{crowding}), or NSGA-III) is better. However, the high HV for the PEARL-NdS (\code{crowding}) in Table \ref{tab:fullmultiobjFOMstwoobj} suggests that the latter approach could provide better volume coverage, but also surpasses the NSGA-based approach on all the other indicators. Moreover, if we only look at the PEARL-NdS (\code{crowding}) versus SO (in the figure and $I_C$ in Table \ref{tab:fullmultiobjFOMstwoobj}), we see that PEARL will propose more Pareto-optimal points. The use of multiple metrics helps therefore to assess the performance of multi-objective approaches \cite{ishibuchi2016sensitivity,seurin2024assessment} and demonstrate PEARL-NdS (\code{crowding}) superiority. Lastly, the use of IGD$^{+}$ on top of IGD did not provide more information, as the solution set of PEARL-NdS (\code{crowding}) comprises most of the Pareto front, hence its IGD was already largely superior to the other algorithms.

More statistical weights was given to $F_{\Delta h}$. The five best Pareto-optimal points are given by $[1.42,527.9]$ found by TS, $[1.43 ,543.2]$ and $[1.44 ,546. ]$ found by NSGA-III, $[1.45 ,549.1]$ found by SA, and $[1.456,553.5]$ found by the PEARL-NdS (\code{crowding})-based approach. Among the PEARL algorithms, this provides an additional reason to consider PEARL-NdS (\code{crowding}) with \code{CL} as potentially superior. Furthermore, the best candidate has a very high $L_C$  compared to the limit of 500 in \cite{seurin2022pwr,seurin2024assessment}.  However, approaching the limit of 1.45 by turning peaking into a constraint rather than an objective could result in lower $L_C$ but also discard the two best PEARL-NdS (\code{crowding}) cases mentioned above. Nevertheless, in Section \ref{sec:constrainedbenchmarkproblems} we will investigate the impact of introducing constraints to guide the search in PEARL for this set of problems and illustrate its potential to still yield high-quality feasible solutions.

For the Three-obj case, most techniques found a Pareto-optimal point (C\_metric $\ne 0.0$) except PEARL-e (\code{cos}). Nevertheless again, the PEARL-NdS (\code{crowding}) approach performs better in terms of HV as seen in Table \ref{tab:fullmultiobjFOMsthreeobj}. The high HV for the PEARL-NdS (\code{crowding}) technique is also illustrated by Figures \ref{fig:fomsunconstrained3d}.(c) to \ref{fig:fomsunconstrained3d}.(e). There, we observe a significant variance in the objective space explored when using the latter method, while PEARL-\code{$\epsilon$} and PEARL-e exacerbate a very low one. Although the $\sigma$ (i.e., extent of the lighter area) over time steps is large, the mean value does not vary substantially and it resulted in solutions that are concentrated around relatively high $L_C$, intermediate $F_{\Delta h}$ and average enrichment. The PEARL-\code{$\epsilon$}-based approach outperforms the other algorithms in terms of GD, $I_{\epsilon}^+$, $I_C$ and C\_metric but exacerbates a very poor HV, IGD, and IGD$^{+}$. The latter has found lots of high-quality points (high $I_C$ equal to 1755 which is the highest) but these points are highly localized, as corroborated by the high IGD. Additionally, The SO-based approach (here NSGA-III) only outperforms PEARL on the IGD$^{+}$. This underscores again the importance of employing multiple multi-objective metrics when analyzing the performance of the algorithms \cite{ishibuchi2016sensitivity,seurin2024assessment}. 

Furthermore, it's worth noting that in PEARL-NdS (\code{crowding}) and PEARL-$\epsilon$, the agents prioritize minimizing low enrichment as their first objective. This choice may be explained by the fact that average enrichment is relatively easier to correlate, as it is derived from altering the composition of the FAs selected to fill a pattern. PEARL-NdS and PEARL-\code{$\epsilon$} are ranking strategies and are seeking solutions of higher rank, ones that dominates others, while PEARL-e is based on the magnitude of the objectives. Looking for lower enrichment will provide the sought after higher ranking. Conversely, achieving higher $L_C$ and lower $F_{\Delta h}$ values is more complex and depends not only on the composition of the chosen FAs but also on the core's arrangement. If we focus on the PEARL-NdS (\code{crowding}) approach, we observe that the algorithm starts to optimize the other two FOMs after about 50 generations or 25,000 samples. Moreover, the $L_C$ increases more smoothly, which might be due to the fact that there is an almost perfect linear relationship between $L_C$ and the average enrichment, hence it is easier to optimize it compared to $F_{\Delta h}$. For high $F_{\Delta h}$, the trade-off with $L_C$ does not fully appear. On top of that, it is very interesting to note that $L_C$ and the average enrichment are nearly perfectly correlated negatively, hence it might be very hard to optimize both at the same time. However, the agents of PEARL-NdS (\code{crowding}) traverse the objective space one objective after the other, resulting in a very high HV compared to the other methods.  
 
 Interestingly again, the preferential set of best solutions per our statistical Shannon metric are all exhibiting low $F_{\Delta h}$ and average enrichment: $[1.43,4.38,501.4]$, $[1.449,4.12,478.9]$, $[1.45,4.18,483.6]$, $[1.44,4.37,507.2]$,$[1.44,4.43,513.3]$. In this instance, all solutions the second  was found by the PEARL-NdS (\code{crowding}) approach and the others by NSGA-III. The first, fourth, and fifth best solutions would meet the feasibility criteria outlined in  \cite{seurin2022pwr,seurin2024assessment}, where our goal was to achieve $L_C \ge 500$ EFPD and $F_{\Delta h} \le 1.45$.
 
\begin{table}[H]
    \centering
        \caption{Different FOMs for the Two-obj case. The nadir is $[2.653,481.2]$ and the utopia is $[1.42,558.0]$.  They are obtained from cases generated over the 6 optimization runs. All the indicators are calculated only with the Pareto equivalent points. The C\_metric is equal to 0 when no solution belongs to the Pareto front. $I_C$ is the number of Pareto-optimal points for the algorithm. The best cases are highlighted in red.}
    \begin{tabular}{c|c|c|c|c|c|c}
    \hline
    FOMs &  PEARL-e (\code{cos}) &  PEARL-\code{$\epsilon$} & PEARL NdS (\code{crowding}) & NSGA-III & SA & TS\\
    \hline
    HV  &   69.805 &  82.337 &  \textcolor{red}{92.054} & 87.786 & 87.623 & 82.922\\
    GD &4.309 & 0.898 & \textcolor{red}{0.032} & 0.752 & 1.562 & 1.562 \\
IGD &11.949 & 4.365 & \textcolor{red}{1.200} &  2.672 & 2.507 & 4.523\\
IGD$^{+}$ & 11.652 & 3.574 & \textcolor{red}{0.003} & 1.882 & 1.572 & 4.403 \\
$I_{\epsilon}^+$ & 0.325 &  0.238 &  \textcolor{red}{0.0170} & 0.0900 & 0.074 & 0.51\\
$I_C$ & 0 & 0 & \textcolor{red}{54} & 2  & 1 & 1\\
C\_metric & 0.0 & 0.0 & \textcolor{red}{0.964} & 0.250 & 0.167 & 0.100\\
\hline
     \hline
    \end{tabular}
    \label{tab:fullmultiobjFOMstwoobj}
\end{table}

 \begin{table}[H]
    \centering
        \caption{Different FOMs for the Three-obj case. The nadir is $[3.606, 4.75,451.5]$ and the utopia $[1.43, 3.911, 553.4]$. They are obtained from cases generated over the 6 optimization runs. All the indicators are calculated only with the Pareto equivalent points. The C\_metric is equal to 0 when no solution belongs to the Pareto front. $I_C$ is the number of Pareto-optimal points for the algorithm. The best cases are highlighted in red.}
    \begin{tabular}{c|c|c|c|c|c|c}
    \hline
    FOMs &  PEARL-e (\code{cos}) &  PEARL-\code{$\epsilon$} & PEARL NdS (\code{crowding}) & NSGA-III & SA & TS\\
    \hline
    HV & 84.883 & 75.703 & \textcolor{red}{98.693} & 97.322 & 89.212& 86.493\\
    GD & 0.129 & \textcolor{red}{0.00332} &  0.0288 & 0.0386  &0.093& 0.107\\
    IGD & 0.621 &  7.156 & \textcolor{red}{0.139}& 0.547 &1.830&1.063\\
    IGD$^{+}$ & 0.129& 2.045& 0.077& \textcolor{red}{0.016}& 0.401& 0.356\\
    $I_{\epsilon}^+$ & 0.178 & \textcolor{red}{0.0699} & 0.122&  0.126  &0.140& 0.080\\
    $I_{C}$ & 0 & \textcolor{red}{1755} & 1152 & 290  &29& 3\\
    C\_metric & 0.0 & \textcolor{red}{0.962} & 0.823 & 0.687  &0.141& 0.007\\
\hline
\end{tabular}
    \label{tab:fullmultiobjFOMsthreeobj}
\end{table}

Additionally, it's noteworthy that in both optimization scenarios, PEARL-e under-performs when compared to the SO-based approaches, even though it performed well in the classical benchmark cases. This difference in performance could be attributed to the arbitrary choice of $\alpha = [1.0, 1.0, 1.0]$. Given that we didn't possess prior knowledge of the Pareto front's shape, we opted for a uniform generation of rays. It is conceivable that a designer could iteratively construct more suitable rays by conducting multiple optimizations in advance. Nevertheless, we recommend exercising caution when considering PEARL-e if no a-priori assumptions about the Pareto front can be made, and PEARL-NdS allows the user to workaround this challenge.

With regards to the SO-based approaches, NSGA-III outperforms the other methods. This was observed in our parallel works with single-objective optimization \cite{seurin2023can,seurin2024surpassing}. There, Evolutionary Strategy (ES), a GA-based approach, outperformed SA and TS. There, our problem requires global search methods to be able to find high quality solutions rapidly, but fail to improve more in the long run because the change in one solution to the next (i.e., the crossovers) were too important. By contrast, local search methods (i.e., similar to SA and TS) only perturb few entries of the input candidate slowing down the progression of the optimization, but enables long-term steady improvements. Nevertheless, the policy in RL, which controls the number of entries of the input space perturbed, manages to be both a global search at the beginning during exploration, while act as a local search in the long-term, enabling fast and constant improvement over time. This behaviors seems to be observed in the multi-objective case as well.

To sum up, for the unconstrained problems with two and three objectives, the PEARL-NdS approach, in particular the \code{crowding}-based one, exacerbates high HV but also $I_C$, and C\_metric surpassing SO systematically, while NSGA-III is the best out of the legacy approaches. PEARL-\code{$\epsilon$} may presents high $I_C$, but the solutions found are usually too localized, which resulted in a lower HV. PEARL-e always under-performed, and tweaking the $\alpha$ might help improving it. However, for real-world application, a practitioner may prefer to limit the amount of tuning, especially for problems encountered for the very first time. For all of these reasons, we will pursue the study with PEARL-NdS and NSGA-III. 

On top of that, for engineering problems, many constraints exist, which is the case for the PWR problem as well \cite{seurin2022pwr,seurin2024assessment}. The next Section \ref{sec:constrainedbenchmarkproblems} consider the same problems as this section but add three constraints namely, $F_q$, $C_b$, and $Bu_{max}$. 

\subsubsection{Constrained Benchmark problems}
\label{sec:constrainedbenchmarkproblems}
\begin{figure}[H]
    \centering
    \includegraphics[scale=0.5]{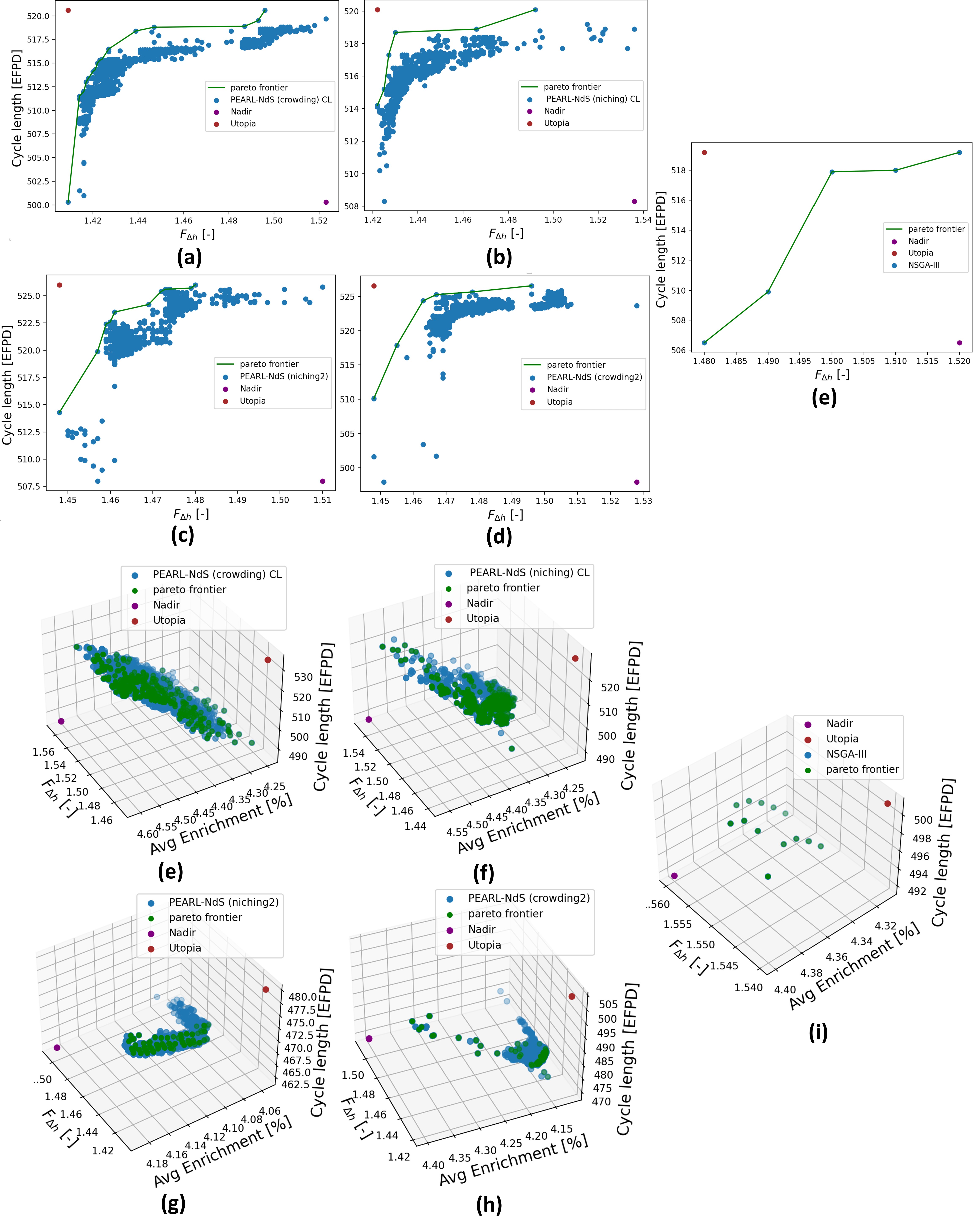}
    \caption{Two-obj ((a) to (d)) and Three-obj ((e) to (i)) feasible solutions generated for all the algorithms utilized. The Pareto front is a line for the Two-obj case, while it is drawn with scattered points for the Three-objective one.}
    \label{fig:bothcsontrained}
\end{figure}

 For the constraints, we consider $F_q \le 1.85$, $C_b \le 1200$ ppm, and Peak Bu $\le 62$ GWd/tHm, similar to \cite{seurin2022pwr,seurin2024assessment}. For the following with omit the "C-" in front of "PEARL" in the text and in the figures. The rank-based method, PEARL-NdS (\code{niching2}) and PEARL-NdS (\code{crowding2}) are using \code{CL} by default, hence we omit to mention it every time. All results with PEARL-NdS cases using \code{CL} will be showcased. Here, 50,000 samples will be collected for both the Two- and Three-obj cases.  We use \code{$\lambda$\_} equal to 32, \code{$\mu$} equal to 2, \code{mutpb} to 0.3, and \code{cxpb} to 0.65 in the both cases. We found that these hyper-parameters provided a good trade-off between exploration and exploitation for the NSGA in each instance to satisfy the constraints and then optimize in the feasible space. Since PEARL will propose more solutions from the different buffers, we will provide the pareto front of all the solutions ever found from NSGA-III rather than the last population (i.e., 32 of them) for fairness. The individual results for the Two- ((a) to (d)) and Three-obj ((e) to (i)) cases are given in Figure \ref{fig:bothcsontrained}. The hyper-parameters utilized are the same as the previous section. We use \code{M} equal to \code{$\kappa$} (= 128).

As can be seen from Figure \ref{fig:bothcsontrained} for the Two-obj case, using PEARL-NdS with \code{CL} outperforms the NSGA-III based approach which could only find 201 feasible solutions. On top of that, having the feasibility in the ranking process (i.e., PEARL-NdS (\code{niching2}) and PEARL-NdS (\code{crowding2}) approaches) resulted in higher $L_C$ but higher $F_{\Delta h}$. 


Moreover, the $L_C$, $C_b$, and $Bu_{max}$ penalties for the distance-based (i.e.,  PEARL-NdS (\code{crowding}) and PEARL-NdS (\code{niching}) approaches) are greater. As a result the number of feasible solutions found is higher for the rank-based approach earlier before the PEARL-NdS (\code{crowding}) catches up  (see top left figure of the leftmost sets of plots). Additionally, because we saw in the previous section that there is a trade-off between $L_C$ and $F_{\Delta h}$ for low values of the latter, when the agents in the distance-based cases start to concentrate on the feasible space (around generation 60), they may uncover solutions with lower $F_{\Delta h}$. Conversely, we see for the PEARL-NdS (\code{niching2}) case, the high $L_C$ comes with the cost of still seeing a non-negligible amount of infeasible solutions (black curve on the top left of the leftmost figure), which is mostly due to exceeding the $C_b$ and more rarely the $Bu_{max}$ limits. 

One hypothesis suggests that many infeasible solutions exhibit similar rewards in the rank-based approaches. This is because numerous different infeasible solutions can be assigned the same rank. As the learning process progresses, it becomes easier to discover feasible solutions that offer higher rewards. Consequently, once the algorithm identifies the feasible region within the objective space, it expeditiously seeks solutions with high rewards (which are solutions with high $L_C$ and low average enrichment in the previous section for the Two- and Three-obj cases, respectively), which often favor the optimization of $L_C$ due to its relative ease compared to the other objective, $F_{\Delta h}$. This phenomenon is not observed in the distance-based approach, where each infeasible solution retains its unique reward and potentially resulted in solutions with lower $F_{\Delta h}$. This distinction may help explain the differences observed in Figure \ref{fig:bothcsontrained}.
 
Among the resulting best solutions, which include $[1.409,500.3]$, $[1.414,511.5]$, $[1.416,512.]$, $[1.417,513]$, and $[1.418,513.4]$ all were found with the PEARL-NdS (\code{crowding}) with \code{CL} method exclusively. Additionally, these solutions conform to the feasibility constraints outlined in \cite{seurin2022pwr,seurin2024assessment}. In Table \ref{tab:fullmultiobjFOMstwoobjconstr}, you can find additional MOO metrics pertaining to the Two-obj case. As previously mentioned, despite the distance-based methods uncovering a smaller proportion of feasible solutions (as indicated in the last row of the Table), they excel in discovering a greater diversity among these solutions. Consequently, the PEARL-NdS (\code{crowding}) and PEARL-NdS (\code{niching}) approaches exhibit higher HV values. The former also boast the highest number of solutions belonging to the Pareto front (as indicated by the $I_C$ in the Table) although \code{niching2} achieve the lowest IGD, GD, and $I_{\epsilon}^+$ values.

For the Three-obj case, it's worth noting that cases PEARL-NdS (\code{niching2}) and PEARL-NdS (\code{crowding2}) tend to discover feasible solutions rapidly, and the transition from finding none to finding feasible solutions is relatively abrupt compared to the distance-based approaches. This behavior may be explained by the fact that PEARL-NdS (\code{niching2}) and  PEARL-NdS (\code{crowding2}) initially select solutions with lower average enrichment after finding solutions, as these are better ranked and allow for the quick discovery of numerous feasible solutions. As mentioned in Section \ref{sec:unconstrainedbenchmarkproblems}, the path of least resistance to find these solutions is to prioritize the optimization of average enrichment first. In contrast, the transitions in the approaches PEARL-NdS (\code{niching}) and PEARL-NdS (\code{crowding}) with \code{CL} are smoother. Attaining higher rewards by moving closer to the feasible space is easier, and there is less need for penalties on enrichment, resulting in solutions with higher $L_C$. In contrast to the Two-obj case however, it may cause lower HV for the latter methods (see the HV for PEARL-NdS (\code{crowding2}) in Table \ref{tab:fullmultiobjFOMsthreeobjconstr}). When considering sample efficiency and employing a multi-metric approach, the distance-based approach seems therefore to be favored again. However, it's worth noting that the preference is not as pronounced as it was for the Two-obj cases.

%

 \begin{table}[H]
    \centering
        \caption{Different FOMs for the Two-obj case. The nadir is $[1.536,497.9]$ and the utopia $[1.409, 526.6]$. They are obtained from cases generated over the 6 optimization runs. The best cases are highlighted in red.}
    \begin{tabular}{c|c|c|c|c}
    \hline
    FOMs &  \makecell{PEARL-NdS with \code{CL} \\ \code{niching}} / \code{crowding} & \makecell{PEARL-NdS \\ \code{niching2} / \code{crowding2}} & NSGA-III & Optimal direction\\
    \hline
    HV &  2.4133 / \textcolor{red}{2.541} & 2.3196  / 2.301&  0.948 & higher\\
    GD & 0.0841 / 0.144  &  \textcolor{red}{0.0228} / 0.334 &  1.664& lower\\
    IGD & 2.846 / 1.784& \textcolor{red}{1.256} / 1.842 & 3.735& lower\\
    IGD$^{+}$ & 1.942 / 1.732 & 0.042 / \textcolor{red}{0.020} & 3.3735 & lower\\
    $I_{\epsilon}^+$ & 0.0350 / 0.046 & \textcolor{red}{0.0270} / 0.039 & 0.083& lower\\
    $I_{C}$ & 2 / \textcolor{red}{11}&  7 / 4 & 0& higher\\
    C\_metric & 0.3333 / \textcolor{red}{0.6875} & 0.7 / 0.667 & 0& higher\\
    Feasible & 27115 / 28656 & 29288 / \textcolor{red}{31039} & 201& higher\\
\hline
\end{tabular}
    \label{tab:fullmultiobjFOMstwoobjconstr}
\end{table}

For the Three-obj case the resulting best solutions are $[1.469,4.05493134,462.2]$, $[1.46,4.06292135,464.6]$, $[1.445,4.07890137,465.5]$, $[1.456,4.07091136,465.5]$, and $[1.457,4.07091136,465.6]$. They were all found with \code{niching2}, which is not surprising since solutions with low enrichment were preferred. 

 \begin{table}[H]
    \centering
        \caption{Different FOMs for the Three-obj case. The nadir is $[1.567,4.608,462.2]$ and the utopia $[1.408,4.055,533.7]$. They are obtained from cases generated over the 6 optimization runs. The best cases are highlighted in red.}
    \begin{tabular}{c|c|c|c|c}
    \hline
    FOMs &  \makecell{PEARL-NdS with \code{CL} \\ \code{niching}}/\code{crowding} & \makecell{PEARL-NdS \\ \code{niching2} / \code{crowding2}} & NSGA-III & Optimal direction\\
    \hline
 HV & 2.134 / 2.000 & 1.347 / \textcolor{red}{2.222} & 0.267 & higher\\
 GD & 0.003 / 0.0207 & \textcolor{red}{0.0} / 0.00725 & 0.0148 & lower\\ 
 IGD & 2.499 / \textcolor{red}{1.878} & 27.425 / 8.714 & 12.776 & lower\\
 IGD$^{+}$ & 0.250 / \textcolor{red}{0.030} & 27.423 / 8.526 & 10.079 & lower \\
  $I_{\epsilon}^+$ & 0.0180  / 0.0570& \textcolor{red}{0.0} / 0.0320 & 0.0774 & lower\\
 $I_{C}$ & \textcolor{red}{330} / 171 & 67 / 40 & 0 & higher\\ 
 C\_metric & 0.948 / 0.584 & \textcolor{red}{1.0} / 0.952 & 0.0 & higher\\ 
 Total feasible & 26699 / 25662 & 30934 / \textcolor{red}{31002} & 157 & higher\\
\hline
\end{tabular}
    \label{tab:fullmultiobjFOMsthreeobjconstr}
\end{table}

For the PEARL-NdS \code{CL}-based approaches, the $L_C$ initially exhibits a slight increase, followed by a decrease up to generations 40-60 to satisfy all constraints, particularly those related to $C_b$ and $Bu_{max}$. Once the agents reach and maintain a position within the feasible region of the objective space $\mathbb{O}$, $L_C$ steadily increases for the Two-obj case because it is easier to optimize than $F_{\Delta h}$. In contrast, for the Three-obj case, the average enrichment first decreases and then only $L_C$ continues to increase. This effect is more pronounced for the rank-based methods, which are seeking high reward rapidly and tend to over-optimize the first objective.  In such cases, applying a distance-based approach to deal with the constraints rather than a rank-based one usually results in solution with either broader volume coverage (for the Two-obj case) or superior average performance across each objective (both Two- and Three-obj cases). It could be interesting to run all cases up to 100,000 samples to see whether the distance-based approaches catches up, but we left that for future research endeavors. 

\subsection{Summary of the results obtained}
\label{sec:summaryoftheresults}

In Section \ref{sec:testingapproaches} we demonstrated the application of PEARL (and its constraints handling counterpart C-PEARL) on classical benchmarks. In Section \ref{sec:applicationofPEARLtopwr} we apply these methods to the classical PWR fuel LP optimization problem in Two- and Three-objective settings. In these scenarios, we demonstrated the potential for PEARL to outperform legacy approaches in a small amount of computing samples, and emphasize the role and performance of the \code{CL} with distance- and rank-based techniques. Tables \ref{tab:summaryofthemethodunconstr} and \ref{tab:summaryofthemethodconstr} summarize the performance of each algorithm studied on the benchmarks and against the NSGA (best algorithm out of the SO-based methods) approach, in the unconstrained and constrained cases, respectively. For each problem we assign "1" if the performance was satisfying, while 0 if the performance was not. For clarity we also added a column "Type for objectives" and "Type for constraints", which emphasize the method the agents are utilizing to traverse the search space. "rank" means that the reward assigned to the solution generated is obtained by ranking the solution within a dynamically updated buffer, while "distance" means that the pure magnitude of an objective is utilized instead. Overall, on all problems and both in the unconstrained and constrained cases, the PEARL-NdS surpasses all the PEARL methods as well as NSGA.

\begin{table}[H]
    \centering
    \caption{Summary of the performance of the methods developed and tested in this paper for the unconstrained cases. The performance depends on the problem studied but here we provide an average.}
    \begin{tabular}{c|c|c|c|c}
    \hline
Algorithms  &  \makecell{Type for \\ objectives} &  \makecell{Classical\\ benchmarks} & \makecell {PWR\\ benchmarks} & \makecell{Performance \\ against NSGA}\\
  \hline
   \makecell{PEARL-NdS\\ (\code{niching})} & rank  & 7/7 & -$^{\star}$ & Surpass$^{\star \star}$ \\
   \makecell{PEARL-NdS\\ (\code{crowding})} & rank & 7/7 & 2/2 & Surpass\\
  \makecell{PEARL-e\\ (\code{cos})} & distance & 7/7 &   0/2$^{\star \star \star}$ &  Inferior\\
  PEARL-\code{$\epsilon$} & rank  & 7/7 & 1/2 & Inferior \\
  \hline
   \end{tabular}
    \label{tab:summaryofthemethodunconstr}
\end{table}
  \begin{tablenotes}
        \item \scriptsize$^{\star}$ Results marked with a "-" were not shown in this paper.
      \item  \scriptsize$^{\star \star}$this assumes that both (\code{niching}) and (\code{crowding}) have similar performance for the PWR cases.
      \item  \scriptsize$^{\star \star \star}$this case could be improved by tweaking the $\alpha$ distribution.
    \end{tablenotes}

\begin{table}[H]
    \centering
    \caption{Summary of the performance of the methods developed and tested in this paper for the constrained cases. The performance depends on the problem studied but here we provide an average.}
    \begin{tabular}{c|c|c|c|c|c}
    \hline
Algorithms  &  \makecell{Type for \\ objectives} & \makecell{Type for \\ constraints} &  \makecell{Classical \\ benchmarks} &  \makecell{PWR \\ benchmarks} & \makecell{Performance \\ against NSGA}\\
  \hline
   \makecell{PEARL-NdS \\ (\code{niching})   \code{CL}}& rank & distance  & 8/8 & 2/2 & Surpass\\
   \makecell{PEARL-NdS \\ (\code{crowding}) \code{CL}}  & rank & distance & 6/6 & 2/2 & Surpass\\
  \makecell{PEARL-NdS \\ (\code{niching2})} & rank & rank  & 8/8 & 2/2 & Surpass\\
    \makecell{PEARL-NdS \\(\code{crowding2})}   &  rank & rank & 8/8 & 2/2 & Surpass\\
 \makecell{ PEARL-e \\ (\code{cos}) \code{CL}} & distance & distance& 8/8 &   - &  -\\
  \makecell{PEARL-\code{$\epsilon$} \\ \code{CL}} & rank & distance & 6/8 & - & - \\
  \hline
  \end{tabular}
    \label{tab:summaryofthemethodconstr}
\end{table}

\section{Concluding remarks}
\label{sec:concludingremarks}
In this study, we devised three variants of the Pareto Envelope Augmented with Reinforcement Learning (PEARL), namely PEARL-e, PEARL-\code{$\epsilon$}, and PEARL-NdS as detailed in Section \ref{sec:unconstrainedPEARL}. These variants were further expanded into C-PEARL in Section \ref{sec:constrainedPEARLcPEARL}, incorporating a constraints satisfaction module. On top of that, thanks to the better performance of PEARL-NdS, we presented results for four sub-versions in Section \ref{sec:constrainedbenchmarkproblems}: PEARL-NdS (\code{crowding}) ,PEARL-NdS (\code{crowding}) in the unconstrained scenario, supplemented by PEARL-NdS (\code{crowding2}) and PEARL-NdS (\code{niching2}) in the constrained scenario. The former leverage a distance-based approach to solve the constraints, while the latter use a magnitude-free approach by ranking solutions.

PEARL is a multi-objective RL-based approach aims at solving engineering design problems with and without constraints. The approach was applied to both classical test sets and PWR problems, demonstrating the potential of PEARL in solving classical and engineering problems with a continuous and combinatorial structure. The approach developed in this paper is a first-of-a-kind that is trying to capture the topological structure of a multi-objective optimization problem within the reward signal sent to the agents. This method differentiates from traditional policy-based multi-objective Reinforcement Learning methods in that it learns a single policy, without the requirement to fall back on multiple neural networks to separately solve simpler sub-problems.

When the Pareto frontier is known (e.g., the dtlz benchmarks studied in Sections \ref{sec:unconstrainedtestsuites} and \ref{sec:constrainedtestsuites}), the envelope version of PEARL, PEARL-e, worked better thanks to the possibility to generate vectors in preferential directions of the objective space. However, when dealing with PWR problems where the front was initially unknown (as discussed in Section \ref{sec:unconstrainedbenchmarkproblems}), the PEARL-NdS-based approaches significantly outperform PEARL-e. Additionally, PEARL-\code{$\epsilon$} has the tendency to discover solutions of good quality, but they often tend to be overly localized. This localization proved advantageous for finding Pareto-optimal solutions in cases like ctp2 and ctp3, which exhibit isolated Pareto-optimal points. However, for our PWR problem, the quality of the HV scores was unsatisfactory. In this context, we applied the original $I_{\epsilon}^{+}$ paradigm, but several emerging methods in Evolutionary Techniques (ET) are actively addressing this issue, including approaches utilizing direction vectors, as discussed in \cite{yang2019ageneralized}. We acknowledge the potential for improving PEARL-\code{$\epsilon$} in future research endeavors,

In the context of the benchmarks, PEARL systematically outperforms TS and SA, but does not always outperform NSGA. However, even if in average the HV could be higher for NSGA in the unconstrained cases (e.g., for dtlz1 and dtlz3 in Section \ref{sec:unconstrainedtestsuites}), there is never a statistically significant advantage. On the other hand, when PEARL is superior in HV, there is such advantage most of the time, especially in the constrained cases. This demonstrate the potential of PEARL to perform well for various types of problem and topologies of the Pareto front.

In the context of the PWR problem, PEARL-NdS consistently outperforms classical SO-based approaches, particularly NSGA, SA, and TS, across various multi-objective metrics, including the HV, as demonstrated in Tables \ref{tab:fullmultiobjFOMstwoobj}, \ref{tab:fullmultiobjFOMstwoobjconstr}, \ref{tab:fullmultiobjFOMsthreeobj}, and \ref{tab:fullmultiobjFOMsthreeobjconstr}. This advantage in HV is particularly noteworthy for core designers, as it enables more efficient exploration of the search space using PEARL-NdS. However, it's important to note that future research should include a more comprehensive statistical analysis to confirm the significant superiority of the newly-developed PEARL approaches, in particular PEARL-NdS, over legacy methods like NSGA, SA, and TS. 

Additionally, we found similar conclusion than observed in our comparison between RL and SO-based approaches in single-objective settings  in another research \cite{seurin2023can,seurin2024surpassing}. There, we demonstrated that the PWR problems are more amenable to global search (i.e., large changes in the input candidates from one step to the next) in the beginning to find high-quality solutions rapidly, and then local search (i.e., small changes in the input candidates from one step to the next) to be able to observe continuous improvement over time. The policy in RL, which adapts over time and hence behaves in this way could explain its superiority in multi-objective optimization as well against NSGA-II and NSGA-III, which are only global-search algorithms, and SA and TS, which are only local-search algorithms. 

Furthermore, we saw that on the PWR problems solved with PEARL-NdS, once the agents reach the trade-off area of the FOMs studied, they will preferentially optimize first the average enrichment, the $L_C$, and $F_{\Delta h}$ by order of simplicity in order to get higher rank and higher reward faster. As a result, even when two objectivse are nearly perfectly negatively correlated (which is the case of the average enrichment and $L_C$), the agents still had the potential to recover a high-quality Pareto front in case of the PEARL-NdS (\code{crowding}) and PEARL-NdS (\code{niching}) versions. However, as discussed in Section \ref{sec:constrainedbenchmarkproblems}, we encountered a potential issue with the rank-based approaches PEARL-NdS (\code{crowding2}) and PEARL-NdS (\code{niching2}) in the constrained case. Once feasible solutions are discovered, the agents tend to prioritize the optimization of a single objective first before the others, resulting in slower learning and reduced sample efficiency. The application of the distance-based approaches PEARL-NdS (\code{crowding}) and PEARL-NdS (\code{niching}) to handle constraints introduces a distinct disparity in rewards between feasible and infeasible solutions. This disparity effectively mitigates the aforementioned effect and broadens the range of solutions found along the Pareto front. To deal with constraints, we separated the problem between finding feasible solutions and finding the optimum Pareto envelope afterward (see Section \ref{sec:constrainedPEARLcPEARL}). There, all constraints were solved at once. A natural enhancement of the method is to create multiple sub-problems combining different constraints together as in \cite{ma2019combinatorial}. This study is the focus of our ongoing efforts.

Lastly, replacing the average enrichment minimization by minimizing the BOC exposure might be a preferable target. Indeed, it compromises with $L_C$ \cite{park2014multicycle} for the current cycle as it reduces the amount of energy extracted from the fuel in the cycle allowing more energy for the following one. The latter presents economic benefits when the end goal is multi-cycle optimization, which is the typical next step in PWR design \cite{mawdsley2022incore,park2014multicycle}. Future research could expand the application of PEARL to such setting.

\section*{Credit Authorship Contribution Statement}

 \textbf{Paul Seurin:} Conceptualization, Methodology, Software, formal analysis \& investigation, Writing– original draft preparation, Visualization. \textbf{Koroush Shirvan:} Investigation, Writing– review \& editing, Supervision, Funding acquisition. All authors have read and agreed to the published version of the  manuscript.
 
 \section*{Data availability}
  PEARL will be a part of the open-source NEORL project (Neuro
Evolution optimization with Reinforcement Learning) \cite{radaideh2023NEORL}, which
 includes a set of implementations of hybrid algorithms combining
 neural networks and evolutionary computation. The documentation for the algorithms in NEORL and examples can be accessed via this link https://neorl.readthedocs.io/en/latest/modules/modules.html.
 \section*{Declaration of Competing Interest}
 The authors declare that they have no known competing financial interests or personal relationships that could have appeared
 to influence the work reported in this paper.

\section*{Acknowledgement}
This work is sponsored by Constellation (formely known as Exelon) Corporation under the award (40008739).

\pagebreak

\newif\ifusebibtex
\usebibtextrue

\ifusebibtex
\setlength{\baselineskip}{12pt}
\bibliographystyle{mc2023}
\bibliography{mc2023}
\else
\setlength{\baselineskip}{12pt}

\fi

\end{document}